%% file: paper.tex
\documentclass{article}
\usepackage{bbm}
\usepackage{comment}
\usepackage[utf8]{inputenc} 
\usepackage[T1]{fontenc}    
\usepackage{url}            
\usepackage{booktabs}       
\usepackage{amsfonts}       
\usepackage{nicefrac}       
\usepackage{microtype}      
\usepackage{amsmath}
\usepackage{amssymb}
\usepackage{graphicx}
\usepackage{subcaption}
\usepackage{natbib}
\usepackage{dsfont}
\usepackage{bm}
\usepackage{dirtytalk}

\usepackage[usenames,dvipsnames]{xcolor}
\usepackage[bookmarks=false]{hyperref}
\hypersetup{
	pdftex,
	pdffitwindow=true,
	pdfstartview={FitH},
	pdfnewwindow=true,
	colorlinks,
	linktocpage=true,
	linkcolor=Green,
	urlcolor=Green,
	citecolor=Green
}

\usepackage{comment}
\usepackage[capitalize,noabbrev]{cleveref}
\usepackage[backgroundcolor=white,textwidth=0.7in]{todonotes}
\DeclareMathOperator*{\argmax}{arg\,max}

\newcommand{\toprank}{\tt TopRank}
\newcommand{\rankoneelim}{\tt Rank1Elim}
\newcommand{\cascadeklucb}{{\tt CascadeKL}$-${\tt UCB}}

\newcommand{\idTSfull}{\tt idTSfull}
\newcommand{\idTS}{\tt idTS}

\newcommand{\idTSvi}{\tt idTSvi}
\newcommand{\idTSivi}{\tt idTSinc}
\newcommand{\ucb}{\tt UCB1}

\usepackage{microtype}
\usepackage{graphicx}
\usepackage{booktabs} 

\usepackage{hyperref}

\newtheorem{theorem}{Theorem}
\newtheorem{assumption}{Assumption}

\usepackage[accepted]{icml2020}

\icmltitlerunning{Influence Diagram Bandits: Variational Thompson Sampling for Structured Bandit Problems}

\begin{document}

\twocolumn[
\icmltitle{Influence Diagram Bandits: Variational Thompson Sampling \\ for Structured Bandit Problems}




\begin{icmlauthorlist}
\icmlauthor{Tong Yu}{to}
\icmlauthor{Branislav Kveton}{goo}
\icmlauthor{Zheng Wen}{deep}
\icmlauthor{Ruiyi Zhang}{ed}
\icmlauthor{Ole J. Mengshoel}{to,ntnu}
\end{icmlauthorlist}

\icmlaffiliation{to}{Carnegie Mellon University}
\icmlaffiliation{goo}{Google Research}
\icmlaffiliation{deep}{DeepMind}
\icmlaffiliation{ed}{Duke University}
\icmlaffiliation{ntnu}{Norwegian University of Science and Technology}

\icmlcorrespondingauthor{Tong Yu}{worktongyu@gmail.com}
\icmlkeywords{}

\vskip 0.3in
]



\printAffiliationsAndNotice{}  

\begin{abstract}
We propose a novel framework for structured bandits, which we call an influence diagram bandit. Our framework captures complex statistical dependencies between actions, latent variables, and observations; and thus unifies and extends many existing models, such as combinatorial semi-bandits, cascading bandits, and low-rank bandits. We develop novel online learning algorithms that learn to act efficiently in our models. The key idea is to track a structured posterior distribution of model parameters, either exactly or approximately. To act, we sample model parameters from their posterior and then use the structure of the influence diagram to find the most optimistic action under the sampled parameters. We empirically evaluate our algorithms in three structured bandit problems, and show that they perform as well as or better than problem-specific state-of-the-art baselines.
\end{abstract}

\input{intro.tex}
\input{background.tex}
\input{setting.tex}
\input{algorithm.tex}
\input{regret_bound.tex}
\input{relatedwork.tex}
\input{experiments.tex}
\input{conclusions.tex}

\clearpage
\bibliography{icml20}
\bibliographystyle{icml2020}

\input{appendix.tex}

\end{document}

%% file: intro.tex
\section{Introduction}

Structured multi-armed bandits, such as combinatorial semi-bandits \citep{gai12combinatorial,chen13combinatorial,pmlr-v38-kveton15}, cascading bandits \citep{kveton2015cascading, li2016contextual}, and low-rank bandits \citep{katariya2017stochastic,bhargava2017active,jun2019bilinear, Lu:2018:EOR:3240323.3240408,zimmert2018factored}, have been extensively studied. Various learning algorithms have been developed, and many of them have either provable regret bounds, good experimental results, or both. Despite such significant progress along this research line, the prior work still suffers from limitations.

A major limitation is that there is no unified framework or general learning algorithms for structured bandits. Most existing algorithms are tailored to a specific structured bandit, and new algorithms are necessary even when the modeling assumptions are slightly modified. For example, cascading bandits assume that if an item in a recommended list is examined but not clicked by a user, the user examines the next item in the list. $\cascadeklucb$ algorithm for cascading bandits \citep{kveton2015cascading} relies heavily on this assumption. In practice though, there might be a small probability that the user skips some items in the list when browsing. If we take this into consideration, $\cascadeklucb$ is no longer guaranteed to be sound and needs to be redesigned. 

\begin{figure*}[t]
  \centering
  \begin{subfigure}[t]{0.23\textwidth}
    \centering
    \includegraphics[width=0.95\textwidth]{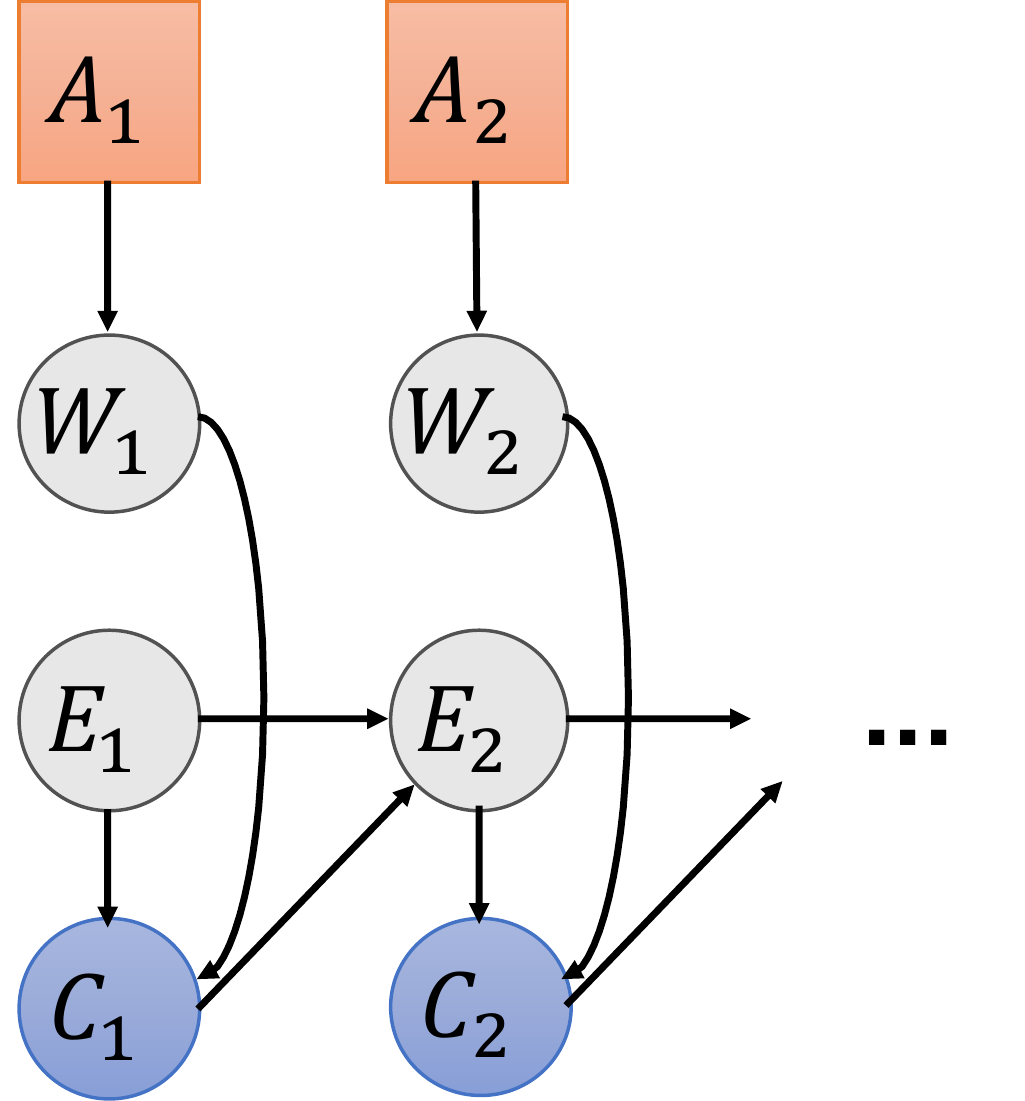}
    \caption{}
    \label{fig:eg1}
  \end{subfigure}
  \vrule
  \begin{subfigure}[t]{0.23\textwidth}
    \centering
    \includegraphics[width=0.9\textwidth]{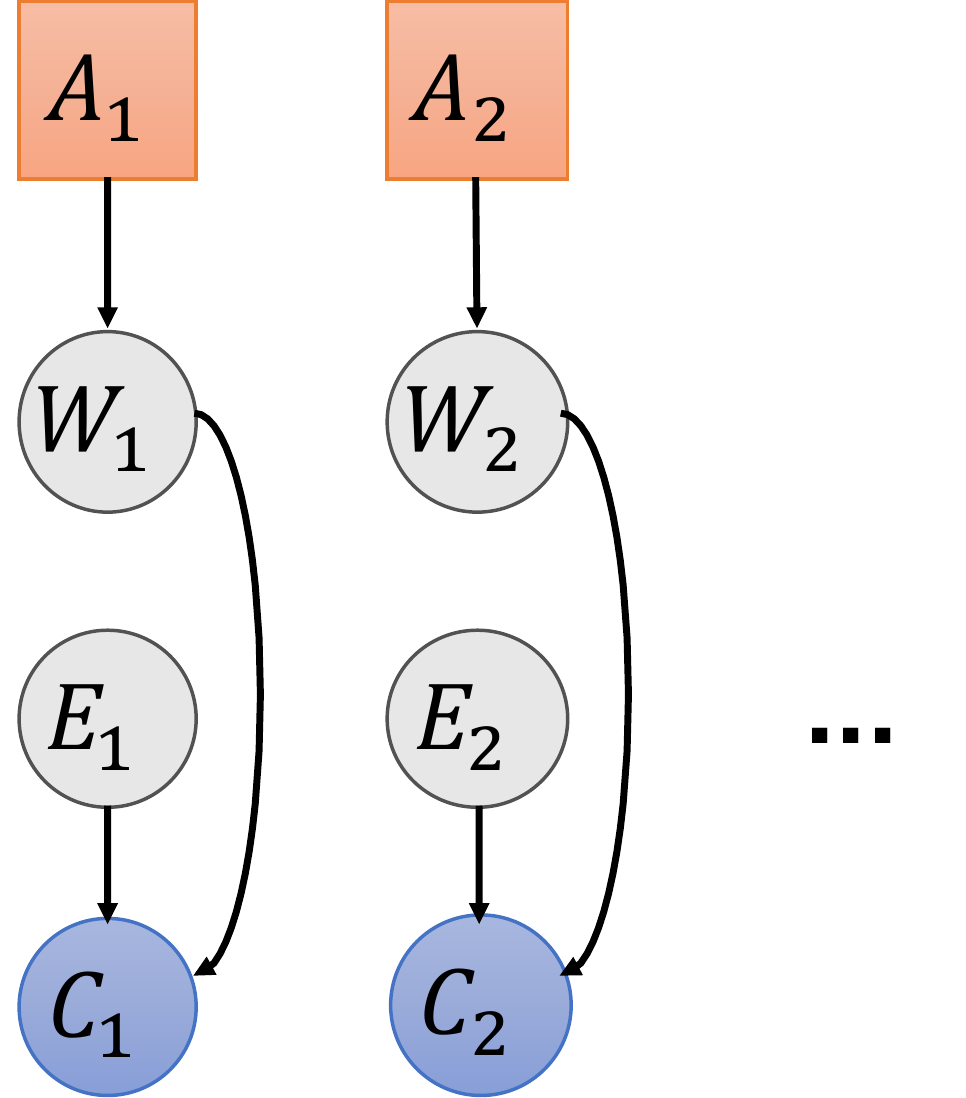}
    \caption{}
    \label{fig:eg2}
  \end{subfigure}
  \vrule
  \begin{subfigure}[t]{0.23\textwidth}
    \centering
    \includegraphics[width=0.85\textwidth]{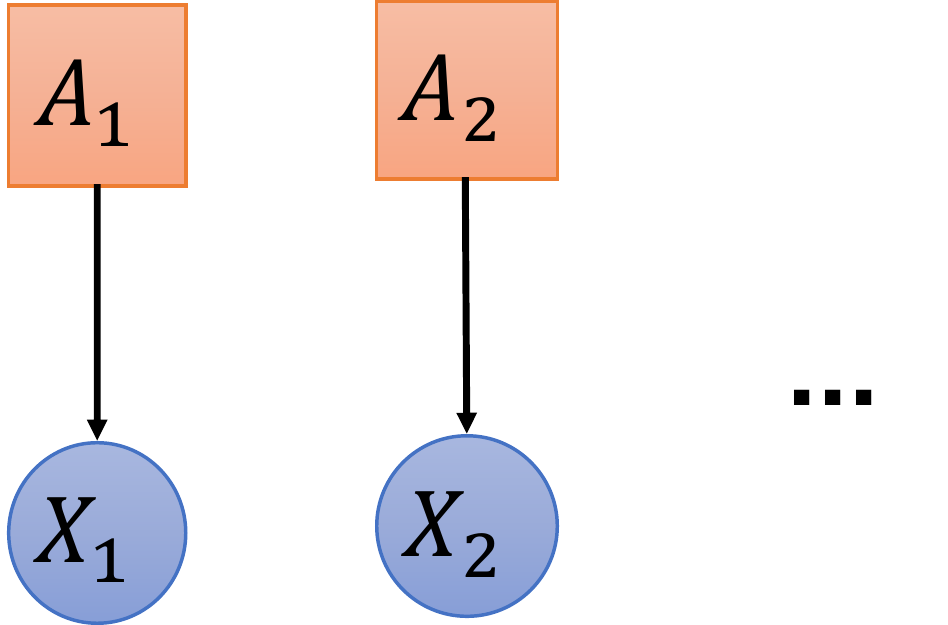}
    \caption{}
    \label{fig:eg3}
  \end{subfigure}
  \vrule
  \begin{subfigure}[t]{0.23\textwidth}
    \centering
    \includegraphics[width=0.55\textwidth]{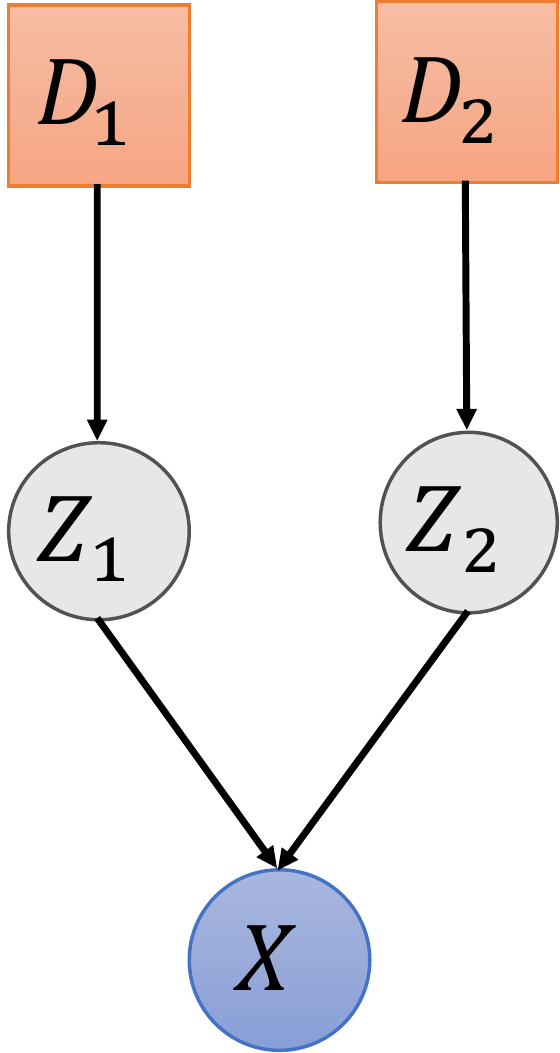}
    \caption{}
    \label{fig:eg4}
  \end{subfigure}
  \caption{Examples of existing models, which are special cases of our framework: (a) cascade model, (b) position-based model, (c) semi-bandit, and (d) rank-$1$ bandit. See the details of the models in Section \ref{sec:setting}. The red nodes are decision nodes, the gray nodes are latent nodes, and the blue nodes are observed nodes. See formal definitions in Section \ref{sec:bg}.
  }
  \label{fig:eg}
\end{figure*}

To enable more general algorithms, we propose a novel online learning framework of influence diagram bandits. The influence diagram \citep{howard1984influence} is a generalization of Bayesian networks. It can naturally represent structured stochastic decision problems, and elegantly model complex statistical dependencies between actions, latent variables, and observations. This paper presents a specific type of influence diagrams (Section \ref{sec:bg}), enabling many structured bandits, such as cascading bandits \citep{kveton2015cascading}, combinatorial semi-bandits \citep{gai12combinatorial,chen13combinatorial,pmlr-v38-kveton15}, and rank-$1$ bandits \citep{katariya2017stochastic}, to be formulated as special cases of influence diagram bandits.

However, it is still non-trivial to efficiently learn to make the best decisions online with convergence guarantee, in an influence diagram with (i) \emph{complex structure} and (ii) \emph{exponentially many actions}. In this paper, we develop a Thompson sampling algorithm $\idTS$ and its approximations for influence diagram bandits. The key idea is to represent the model in a compact way and track a structured posterior distribution of model parameters. We sample model parameters from their posterior and then use the structure of the influence diagram to find the most optimistic action under the sampled parameters. In complex influence diagrams, latent variables naturally occur. In this case, the model posterior is intractable and exact posterior sampling is infeasible. To handle such problems, we propose variational Thompson sampling algorithms, $\idTSvi$ and $\idTSivi$. By compact model representation and efficient computation of the best action under a sampled model via dynamic programming, our algorithms are efficient both statistically and computationally. We further derive an upper bound on the regret of $\idTS$ under additional assumptions, and show that the regret only depends on the model parameterization and is independent of the number of actions.

This paper makes four major contributions. First, we propose a novel framework called influence diagram bandit, which unifies and extends many existing structured bandits. Second, we develop algorithms to efficiently learn to make the best decisions under the influence diagrams with complex structures and exponentially many actions. We further derive an upper bound on the regret of our $\idTS$ algorithm. Third, by tracking a structured posterior distribution of model parameters, our algorithms naturally handle complex problems with latent variables. Finally, we validate our algorithms on three structured bandit problems. Empirically, the average cumulative reward of our algorithms is up to three times the reward of the baseline algorithms.

%% file: background.tex
\section{Background}
\label{sec:bg}

An \emph{influence diagram} is a Bayesian network augmented with decision nodes and a reward function \citep{howard1984influence}. The structure of an influence diagram is determined by a \emph{directed acyclic graph (DAG)} $G$. In our problem, nodes in the influence diagram can be partially observed and classified into three categories: \emph{decision nodes} $A = (A_i)_{i \in \mathbb{N}}$, \emph{latent nodes} $Z = (Z_i)_{i \in \mathbb{N}}$, and \emph{observed nodes} $X = (X_i)_{i \in \mathbb{N}}$. Among these nodes, $X$ are stochastic and observed, $Z$ are stochastic and unobserved, and $A$ are non-stochastic. Except for the decision nodes, each node corresponds to a random variable. Without loss of generality, we assume that all random variables are Bernoulli\footnote{To handle non-binary categorical variables, we can extend our algorithms (\cref{sec:algorithm}) by replacing Beta with Dirichlet.}.
The decision nodes represent decisions that are under the full control of the decision maker. We further assume that each decision node has a single child and that each child of a decision node has a single parent. The edges in the diagram represent probabilistic dependencies between variables. The reward function $r(X, Z)$ is a deterministic function of the values of nodes $X$ and $Z$. In an influence diagram, the \emph{solution} to the decision problem is an instantiation of decision nodes that maximizes the reward.

\subsection{Influence Diagrams for Bandit Problems}

\newcommand{\parent}{\mathsf{par}}
\newcommand{\child}{\mathsf{child}}
\newcommand{\cA}{\mathcal{A}}

The decision nodes affect how random variables $X$ and $Z$ are realized, which in turn determine the reward $r(X, Z)$. To explain this process, we introduce some notation. Let $V$ be a node, $\child(V)$ be the set of its children, and $\parent(V)$ be the set of its parents. For simplicity of exposition, each decision node $A_i$ is a categorical variable that can take on $L$ values $\cA = \{1, \dots L\}$. Each value $a \in \cA$ corresponds to a decision and is associated with a fixed Bernoulli mean $\mu_a \in [0, 1]$. The number of decision nodes is $K$. The \emph{action} $a = (a_1, \dots, a_K) \in \cA^K$ is a tuple of all $K$ decisions.

The nodes in the influence diagram are instantiated by the following stochastic process. First, all decision nodes $A_i$ are assigned decisions by the decision maker. Let $A_i = a_i$ for all $i \in [K]$. Then the value of each $\child(A_i)$ is drawn according to $a_i$. Specifically, it is sampled from a Bernoulli distribution with mean $\mu_{a_i}$, $\child(A_i) \sim \mathrm{Ber}(\mu_{a_i})$. All remaining nodes are set as follows. For any node $V$ that has not been set yet, if all nodes in $\parent(V)$ have been instantiated, $V \sim \mathrm{Ber}(\mu)$, where $\mu \in [0, 1]$ depends only on the assigned values to $\parent(V)$. Since the influence diagram is a DAG, this process is well defined and guaranteed to instantiate all nodes. The corresponding reward is $r(X, Z)$.

The model can be parameterized by a vector of Bernoulli means $\theta \in [0, 1]^{d + L}$. The last $L$ entries of $\theta$ correspond to $\mu_a$, one for each decision $a \in \cA$. The first $d$ entries parameterize the conditional distributions of all nodes that are not in $\cup_{i \in [K]} \child(A_i)$, directly affected by the decision nodes. We denote the joint probability distribution of $X$ and $Z$ conditioned on action $a \in \cA^K$ by $P(X, Z \mid \theta, a)$.

\subsection{Simple Example}

We show a simple influence diagram in Figure \ref{fig:eg4}. The decisions nodes are $A = (A_1, A_2)$, the observed node is $X$, and the latent nodes are $Z = (Z_1, Z_2)$. After the decision node $A_i$ is set to $a_i$, it determines the mean of $Z_i$, $\mu_{a_i}$. Then, after both $Z_i \sim \mathrm{Ber}(\mu_{a_i})$ are drawn, $X$ is drawn from a Bernoulli distribution with mean $P(X = 1 \mid Z_1, Z_2)$.

Since the number of decisions is $L$, this model has $L + 4$ parameters. The $L$ parameters are $(\mu_a)_{a \in \cA}$. The remaining $4$ parameters are $P(X = 1 \mid Z_1, Z_2)$ for $Z_1, Z_2 \in \{0, 1\}$.

%% file: setting.tex
\section{Influence Diagram Bandits}
\label{sec:setting}

Consider an influence diagram with observed nodes $X$, latent nodes $Z$, and decision nodes $A$ in Section \ref{sec:bg}. Recall that all random variables associated with these nodes are binary, and the model is parameterized by a vector of $d+L$ Bernoulli means $\theta_\ast \in [0, 1]^{d+L}$. The learning agent knows the structure of the influence diagram, but does not know the marginal and conditional distributions in it. That is, the agent does not know $\theta_\ast$. Let $r(x, z)$ be the reward associated with $X = x$ and $Z = z$. To simplify exposition, let $r(a, \theta)$ be the expected reward of action $a$ under model parameters $\theta \in [0, 1]^{d+L}$,
\begin{align*}
  r(a, \theta) = \sum_{x, z} r(x, z) P(x, z \mid \theta, a).
\end{align*}
At time $t$, the agent adaptively chooses action $a_t$ based on the past observations. Then the binary values of the children of decision nodes $A$ are generated from their respective Bernoulli distributions, which are specified by $a_t$. Analogously, the values of all other nodes are generated based on their respective marginal and conditional distributions. Let $x_t$ and $z_t$ denote the values of $X$ and $Z$, respectively, at time $t$. At the end of time $t$, the agent observes $x_t$ and receives stochastic reward $r(x_t, z_t)$.

The agent’s policy is evaluated by its $n$-step expected cumulative regret
\begin{align*}
  \textstyle
  R(n, \theta_\ast)
  = \mathbb{E}\left[{\sum_{t=1}^{n}R(a_t, \theta_\ast)} \middle | \theta_\ast \right],
\end{align*}
where $R(a_t, \theta_\ast) = r(a^\ast, \theta_\ast) - r(a_t, \theta_\ast)$ is the instantaneous regret at time $t$, and
\begin{align*}
  \textstyle
  a^\ast = \argmax_{a \in \cA^K}r(a, \theta_\ast)
\end{align*}
is the optimal action under true model parameters $\theta_\ast$. For simplicity of exposition, we assume that $a^\ast$ is unique. When we have a prior over $\theta_\ast$, an alternative performance metric is the $n$-step Bayes regret, which is defined as
\begin{align*}
  R_B (n)
  = \mathbb{E} \left[R(n, \theta_\ast) \right],
\end{align*}
where the expectation is over $\theta_\ast$ under its prior. 

We review several examples of prior works, which can be viewed as special cases of influence diagram bandits, below.

\subsection{Online Learning to Rank in Click Models}
\label{sec:model1}

The \emph{cascade model} is a popular model in \emph{learning to rank} \citep{chuklin2015click}, which has been studied extensively in the bandit setting, starting with \citet{kveton2015cascading}. The model describes how a user interacts with a list of items $a = (a_1, \dots, a_K)$ at $K$ positions. We visualize it in \cref{fig:eg1}. For each position $k \in [K]$, the model has four nodes: decision node $A_k$, which is set to the item at position $k$, $a_k$; latent attraction node $W_k$, which is the attraction indicator of item $a_k$; latent examination node $E_k$, which indicates that position $k$ is examined; and observed click node $C_k$, which indicates that item $a_k$ is clicked. The attraction of item $a_k$ is a Bernoulli random variable with mean $\mu_{a_k}$. Therefore, $P(W_k = 1 \mid A_k = a_k) = \mu_{a_k}$, as in our model. The rest of the dynamics, that the item is clicked only if it is attractive and its position is examined, and that the examination of items stops upon a click, is encoded as
\begin{align*}
  P(C_k = 1 \mid W_k, E_k)
  & = W_k E_k\,, \\
  P(E_k = 1 \mid C_{k - 1}, E_{k - 1})
  & = (1 - C_{k - 1}) E_{k - 1}\,.
\end{align*}
The first position is always examined, and therefore we have $P(E_1 = 1) = 1$. The reward is the total number of clicks, $r(X, Z) = \sum_{k = 1}^K C_k$.

The \emph{position-based model} \citep{chuklin2015click} in \cref{fig:eg2} is another popular click model, which was studied in the bandit setting by \citet{lagree2016multiple}. The difference from the cascade model is that the examination indicator of position $k$, $E_k$, is an independent random variable.

\subsection{Combinatorial Semi-Bandits}
\label{sec:model2}

The third example is a combinatorial semi-bandit \citep{gai12combinatorial,chen13combinatorial,pmlr-v38-kveton15,wen2015efficient}. In this model, the agent chooses $K$ items $a = (a_1, \dots, a_K)$ and observes their rewards. We visualize this model in \cref{fig:eg3}. For the $k$-th item, the model has two nodes: decision node $A_k$, which is set to the $k$-th chosen item $a_k$; and observed reward node $X_k$, which is the reward of item $a_k$. If the reward of item $a_k$ is a Bernoulli random variable with mean $\mu_{a_k}$, $P(X_k = 1 \mid A_k = a_k) = \mu_{a_k}$, as in our model. The reward is the sum of individual item rewards, $r(X, Z) = \sum_{k = 1}^K X_k$.

\subsection{Bernoulli Rank-$1$ Bandits}
\label{sec:model3}

The last example is a Bernoulli rank-$1$ bandit \citep{katariya2017stochastic}. In this model, the agent selects the row and column of a rank-$1$ matrix, and observes a stochastic reward of the product of their latent factors. We visualize this model in \cref{fig:eg4} and can represent it as follows. We have two decision nodes, $A_1$ for rows and $A_2$ for columns. The values of these nodes, $a_1$ and $a_2$, are the chosen rows and columns, respectively. We have two latent nodes, $Z_1$ for rows and $Z_2$ for columns. For each, $P(Z_k = 1 \mid A_k = a_k) = \mu_{a_k}$, where $\mu_{a_k}$ is the latent factor corresponding to choice $a_k$. Finally, we have one observed reward node $X$ such that $P(X = 1 \mid Z_1, Z_2) = Z_1 Z_2$. The reward is $r(X, Z) = X$.

%% file: algorithm.tex
\section{Algorithm}
\label{sec:algorithm}

There are two major challenges in developing efficient online learning algorithm for influence diagram bandits. First, with exponentially many actions, it is challenging to develop a sample efficient algorithm to learn a generalizable model statistically efficiently. Second, it is computationally expensive to compute the most valuable action when instantiating the decision nodes in each step, given exponentially many combinations of options. 

Many exploration strategies exist in the online setting, such as the $\epsilon$-greedy policy \citep{sutton2018reinforcement}, $\ucb$ \citep{auer2002finite}, and Thompson sampling \cite{thompson1933likelihood}. While we do not preclude that UCB-like algorithms can be developed for our problem, we argue that they are unnatural. Roughly speaking, the upper confidence bound (UCB) is the highest value of any action under any plausible model parameters, given the history. It is unclear how to solve this problem efficiently in influence diagrams. In contrast, Thompson sampling is more natural. The model parameters can be sampled from their posterior, and the problem of finding the most valuable action given fixed model parameters can be solved using dynamic programming \citep{tatman1990dynamic}.

\subsection{Thompson Sampling}
\label{sec:learninginfluence}

Thompson sampling \cite{thompson1933likelihood} is a popular online learning algorithm, which we adapt to our setting as follows. Let $\tilde{x} = (x_\ell)_{\ell=1}^{t-1}$, $\tilde{z} = (z_\ell)_{\ell=1}^{t-1}$, and $\tilde{a} = (a_\ell)_{\ell=1}^{t-1}$ be the values of observed nodes, latent nodes, and decision nodes, respectively, up to the end of time $t - 1$. First, the algorithm samples model parameters $\theta_t \sim p_{t-1}$, where $p_{t-1}$ is the posterior of $\theta_\ast$ at the end of time $t-1$. That is,
\begin{align*}
  p_{t-1}(\theta) = P(\theta_\ast = \theta \mid \tilde{x}, \tilde{a})
\end{align*}
for all $\theta$. Second, the action at time $t$ is chosen greedily with respect to $\theta_t$,
\begin{align*}
  \textstyle
  a_t = \argmax_{a \in \cA^K}r(a, \theta_t).
\end{align*}
Finally, the agent observes $x_t$ and receives reward $r(x_t, z_t )$. We call this algorithm $\idTS$. Note that $\idTS$ is generally computationally intractable when latent nodes are present, due to the need to sample from the exact posterior.

\subsection{Fully-Observable Case}

In the fully-observable case, with no latent variables and Beta prior, $\idTS$ is computationally efficient. To see it, note that in this case $p_{t-1}$ factors over model parameters, and is a product of Beta distributions. Based on the observed node values, we can update the Beta posterior for each model parameter in $\theta_\ast$ individually and computationally efficiently, since the Beta distribution is the conjugate prior of the Bernoulli distribution. One example of this case is the combinatorial semi-bandit in Section \ref{sec:model2}.

\subsection{Partially-Observable Case}
\label{sec:realistic}

In complex influence diagram bandits, such as rank-$1$ bandits, latent variables are typically present. In such cases, exact sampling from $p_{t-1}$ is usually computationally intractable, which limits the use of $\idTS$. However, there are many computationally tractable approximations to sampling from $p_{t-1}$, such as variational inference and particle filtering \cite{bishop2006pattern,andrieu2003introduction}. In practice, the performance of particle filtering heavily depends on different settings of the algorithm (\emph{e.g.}, number of particles and transition prior). Thus, we develop an approximate Thompson sampling algorithm, $\idTSvi$, based on variational inference. We compare our algorithms to particle filtering later in Section \ref{sec:com2pf}.

To keep notation uncluttered, we omit $\tilde{a}$ below. Let $q(\tilde{z}, \theta)$ be a factored distribution over $(\tilde{z}, \theta)$. To approximate the posterior $p_{t-1}$, we approximate $P(\tilde{z},\theta\mid \tilde{x})$ by minimizing its difference to $q(\tilde{z}, \theta)$. To achieve this, we decompose the following log marginal probability using
\begin{align*}
	\log P(\tilde{x}) = & \int_{\theta} \sum_{\tilde{z}}q(\tilde{z}, \theta) \log P(\tilde{x}) d\theta  \nonumber\\
	= &   \int_{\theta} \sum_{\tilde{z}}q(\tilde{z}, \theta) \log \frac{P(\tilde{z},\theta\mid \tilde{x})P(\tilde{x})q(\tilde{z},\theta)}{P(\tilde{z},\theta\mid \tilde{x})q(\tilde{z},\theta)} d\theta \nonumber\\
	= & \int_{\theta} \sum_{\tilde{z}}q(\tilde{z}, \theta) \log \frac{P(\tilde{x},\tilde{z},\theta)}{q(\tilde{z},\theta)} d\theta  \nonumber\\
	& + \int_{\theta} \sum_{\tilde{z}}q(\tilde{z}, \theta) \log \frac{q(\tilde{z},\theta)}{P(\tilde{z},\theta\mid \tilde{x})} d\theta.
\end{align*}
We denote the first term by $\mathcal{L}(q)$. The second term is the KL divergence between $P(\tilde{z},\theta\mid \tilde{x})$ and $q(\tilde{z}, \theta)$. Note $P(\tilde{x})$ is fixed and the KL divergence is non-negative. Therefore, we can minimize the KL divergence by maximizing $\mathcal{L}(q)$ with respect to $q$, to achieve better posterior approximations. 

We maximize $\mathcal{L}(q)$ as follows. By the mean field approximation, let the approximate posterior factor as
\begin{align}
\label{eq:qzt}
  q(\tilde{z}, \theta) = q(\theta) \Pi_{\ell = 1}^{t-1} q_{\ell} (z_{\ell}),
\end{align}
where $q(\theta)$ is the probability of model parameters $\theta$ and $q_\ell(z_\ell)$ is the probability that the values of latent nodes at time $\ell$ are $z_\ell$. Since $\theta_i$ is the mean of a Bernoulli variable, we factor $q(\theta)$ as $\Pi_{i = 1}^{d+L} \theta_i^{\alpha_i} (1-\theta_i)^{\beta_i}$ and represent it as $d + L$ tuples $\{(\alpha_i, \beta_i)\}_{i = 1}^{d + L}$. For any $\ell$, $q_\ell(z_\ell)$ is a categorical distribution. From the definition of $P(x, z, \theta)$, we have 
\begin{align}
\label{eq:pxzt}
  \log P(\tilde{x},\tilde{z},\theta) = \log P(\theta) + \sum_{\ell = 1}^{t-1} \log P(x_\ell, z_\ell\mid \theta).
\end{align}
By combining (\ref{eq:qzt}) and (\ref{eq:pxzt}) with the definition of $\mathcal{L}(q)$, we get that
\begin{align}
\label{eq:lq}
	  \mathcal{L}(q) = & 
	 \int_\theta \sum_{\tilde{z}} q(\tilde{z}, \theta) \log P(\tilde{x}, \tilde{z}, \theta) d \theta \nonumber \\
	& - \int_\theta \sum_{\tilde{z}} q(\tilde{z}, \theta) \log q (\tilde{z}, \theta) d \theta \nonumber 
	\nonumber \\
	 = &  \sum_{\ell = 1} ^{t-1} \int_\theta q(\theta) \sum_{z_\ell} q_\ell(z_\ell) \log P(x_\ell, z_\ell\mid  \theta) d \theta  \nonumber \\ & + \int_\theta q(\theta) \log P(\theta) d \theta   \nonumber - \sum_{\ell = 1}^{t-1}\sum_{z_\ell} q_\ell(z_\ell) \log q_\ell(z_\ell) \nonumber \\ &- \int_\theta q(\theta) \log q(\theta) d \theta
 \end{align}
The above decomposition suggests the following EM-like algorithm \cite{dempster1977maximum}. 

First, in the E-step, we fix $q(\theta)$. Then
\begin{align*}
	\mathcal{L}(q) = &\sum_{\ell}^{t-1} \sum_{z_\ell} q_\ell(z_\ell) \int_{\theta} q(\theta) \log P(x_\ell, z_\ell\mid  \theta) d \theta  \\ &- \sum_{\ell}^{t-1} \sum_{z_\ell} q_\ell(z_\ell) \log q_\ell(z_\ell)  + C
\end{align*}
for some constant $C$. By taking its derivative with respect to $q_\ell(z_\ell)$ and setting it to zero, $\mathcal{L}(q)$ is maximized by
\begin{align}
  \label{eq:estep}
  q_\ell(z_\ell) \propto \exp \left[ \int_{\theta} q(\theta) \log P(x_\ell, z_\ell\mid  \theta) d\theta \right].
\end{align}
Second, in the M-step, we fix all $q_\ell(z_\ell)$. Then, for some constant $C$,
\begin{align*}
\mathcal{L}(q) 	 = & \int_\theta q(\theta) \sum_{\ell = 1}^{t-1} \sum_{z_\ell} q_\ell(z_\ell) \log P(x_\ell, z_\ell\mid \theta)  d \theta \\
	& + \int_\theta q(\theta) \log P(\theta) d \theta  - \int_\theta q(\theta) \log q(\theta) d \theta + C.
\end{align*}
By taking its derivative with respect to $q(\theta)$ and setting it to zero, $\mathcal{L}(q)$ is maximized by
\begin{align}
\label{eq:mstep}
q(\theta) \propto \exp \left[\log P(\theta) + \sum_{\ell = 1}^{t-1}\sum_{z_\ell} q_\ell(z_\ell) \log P(x_\ell, z_\ell\mid \theta)	\right].
\end{align}
The above two steps, the E-step and M-step, are alternated until convergence. This is guaranteed under our model assumption.

The pseudocode of $\idTSvi$ is in \cref{alg:vits}. In line $4$, $\theta_t$ is sampled from its posterior. In line $5$, the agent takes action $a_t$ that maximizes the expected reward under model parameters $\theta_t$. In line $6$, the agent observes $x_t$ and receives reward. From line $7$ to line $16$, we update the posterior of $\theta$, by alternating the estimations of $q_\ell(z_\ell)$ and $q(\theta)$ until convergence. In line $12$, we update $q_\ell(z_\ell)$ by the E-step. In line $14$, we update $q(\theta)$ by the M-step.

\begin{algorithm}[t]
	\caption{$\idTSvi$: Influence diagram TS with variational inference.}
	\label{alg:vits}
	
	\begin{algorithmic}[1]
		\STATE {\bfseries Input:} $\epsilon > 0$
		\STATE Randomly initialize $q$ 
		\FOR{$t = 1, \dots, n$}
		\STATE Sample $\theta_t$ proportionally to $q(\theta_t)$
		\STATE Take action $a_t = \argmax_{a \in \cA^K} r(a, \theta_t)$
		
		\STATE Observes $x_t$ and receive reward $r(x_t, z_t)$
		
		\STATE Randomly initialize $q$
		
		\STATE Calculate $\mathcal{L}(q)$ using (\ref{eq:lq}) and set $\mathcal{L}'(q) = - \infty$
		\WHILE{$\mathcal{L}(q) - \mathcal{L}'(q) \geq \epsilon$}
	\STATE 	Set $\mathcal{L}'(q) = \mathcal{L}(q)$
		\FOR{$\ell = 1, \dots, t$}%
		\STATE  Update $q_\ell (z_\ell)$ using (\ref{eq:estep}), for all $z_\ell$
		\ENDFOR
	\STATE 	Update $q(\theta)$ using (\ref{eq:mstep})
	\STATE 	Update $\mathcal{L}(q)$ using (\ref{eq:lq})
		\ENDWHILE
		\ENDFOR
	\end{algorithmic}
\end{algorithm}
\vspace{0.5cm}

\subsection{Improving Computational Efficiency}
\label{sec:idtsinc}

To make $\idTSvi$ computationally efficient, we propose its incremental variant, $\idTSivi$. The main problem in \cref{alg:vits} is that each $q_t(z_t)$ is re-estimated at all times from time $t$ to time $n$. To reduce the computational complexity of \cref{alg:vits}, we estimate $q_t(z_t)$ only at time $t$. That is, the \say{for} loop in line $11$ is only run for $\ell = t$; and $q_\ell(z_\ell)$ for $\ell < t$ are reused from the past. The pseudocode of $\idTSivi$ is in \cref{sec:idTSivi_app}.

%% file: regret_bound.tex
\newcommand{\E}{\mathbb{E}}
\newcommand{\Omax}{O_{\max}}
\newcommand{\Hist}{\mathcal{H}}

\section{Regret Bound}
\label{sec:regret_bound}

In this section, we derive an upper bound on the $n$-step Bayes regret $R_B(n)$ for $\idTS$ in influence diagram bandits. 

We introduce notations before proceeding. We say a node in an influence diagram bandit is \emph{de facto} observed at time $t$ if its realization is either observed or can be exactly inferred\footnote{Exact inference is often possible when some conditional distributions are known and deterministic. For instance, in the cascade model (\cref{sec:model1}), the attractions of items above the click position can be exactly inferred.} at that time. Let $\theta^{(i)}_\ast$ denote the $i$th element of $\theta_\ast$, which parameterizes a (conditional) Bernoulli distribution at node $j_i$ in the influence diagram. Note that if node $j_i$ has parents $\parent(j_i)$, then $\theta^{(i)}_\ast$ corresponds to a particular realization at nodes $\parent(j_i)$. We define the event $E_t^{(i)}$ as the event that (1) both node $j_i$ and its parents $\parent(j_i)$ (if any) are \emph{de facto} observed at time $t$, (2) the realization at $\parent(j_i)$ corresponds to $\theta^{(i)}_\ast$, and (3) the realization at $j_i$ is conditionally independently drawn from $\mathrm{Ber}\left( \theta^{(i)}_\ast \right)$. Note that under event $E_t^{(i)}$, the agent observes and knows that it observes a Bernoulli sample corresponding to $\theta^{(i)}_*$ at time $t$. To simplify the exposition, if clear from context, we omit the subscript $t$ of $E_t^{(i)}$. 

Our assumptions are stated below.

\begin{assumption}
\label{ass:mono} The parameter index set $ \left \{ 1, \ldots, d+L \right \}$ is partitioned into two disjoint subsets $\mathcal{I}^+$ and $\mathcal{I}^-$. For all $ i \in \mathcal{I}^+$ (or $ i \in \mathcal{I}^-$), $r(a, \theta)$ is weakly increasing (or weakly decreasing) in $\theta^{(i)}$ for any action $a$ and probability measure $\theta \in [0, 1]^{d+L}$.
\end{assumption}

\begin{assumption}
\label{ass:decomp} For any action $a$ and any Bernoulli probability measures $\theta_1, \theta_2 \in [0, 1]^{d+L}$, we have
\begin{small}
\begin{equation*}
     \left | r(a, \theta_1) - r(a, \theta_2) \right| \leq 
    C  
    \sum_{i=1}^{d+L} P(E^{(i)} | \theta_2, a) \left |  \theta_1^{(i)} - \theta_2^{(i)} \right |, 
 \end{equation*}
\end{small}
where $C \geq 0$ is a constant.
\end{assumption}

\cref{ass:mono} says that $r(a, \theta)$ is element-wise monotone (either weakly increasing or decreasing) in $\theta$. \cref{ass:decomp} says that the expected reward difference is bounded by a weighted sum of the probability measure difference, with weights proportional to the observation probabilities. Note that the satisfaction of Assumption~\ref{ass:decomp} depends on both the functional form of $r(X, Z)$ and the information feedback structure in the influence diagram. 

Intuitively, both combinatorial semi-bandits and cascading bandits discussed in \cref{sec:setting} satisfy Assumptions \ref{ass:mono} and \ref{ass:decomp}. The expected reward increases with $\theta$ in both models. Thus Assumption 1 holds. Assumption 2 holds in combinatorial semi-bandits, since all nodes are observed and the expected reward is linear in $\theta$. Assumption 2 holds in cascading bandits due to Lemma 1 in \citet{kveton2015cascading}. The proof is in \cref{sec:assumptions}. 

Before we present our regret bound, we define the metric of \emph{maximum expected observations}
\begin{equation}
\textstyle
\Omax = \max_a \E \left[\sum_{i=1}^{d+L} 
\mathbf{1}\left[E^{(i)} \right] \middle | a \right], \label{eq:omax}
\end{equation}
where the expectation is over both $\theta^*$ under the prior, and the random samples in the influence diagram under $\theta^*$ and $a$. Notice that for any action $a$, $\E \left[\sum_{i=1}^{d+L} \mathbf{1}\left[E^{(i)} \right] \middle | a \right]$ is the expected number of observations under $a$; hence, $\Omax$ is the maximum expected number of observations over all actions. Let $|X|$ and $|Z|$ be the number of observed nodes and latent nodes, respectively. Notice that by definition, $\Omax \leq |X| + |Z|$. Moreover, $\Omax \leq |X|$ if no latent variables are exactly inferred\footnote{Note that a parent of an observed node might not be observed, thus in general $\Omax \leq |X|$.}; 
and $\Omax = |X|$ in the fully observable case.

Our main result is stated below.

\begin{theorem}
\label{thm:regret_bound}
Under Assumptions \ref{ass:mono} and \ref{ass:decomp}, if we apply exact Thompson sampling $\idTS$ to influence diagram bandits,
\begin{align*}
    R_B(n) \leq & \, \mathcal{O} \left( C \sqrt{(L+d) \Omax n }\log n \right),
\end{align*}
where $\Omax$ is defined in \eqref{eq:omax}.
\end{theorem}
Please refer to Appendix~\ref{sec:proof} for the proof of Theorem~\ref{thm:regret_bound}. It is natural for the regret bound to be $\mathcal{O}(\sqrt{\Omax})$. One can see this by considering special cases. For example, in combinatorial semi-bandits, $\Omax=K$ and \citet{pmlr-v38-kveton15} proved a $\mathcal{O}(\sqrt{K})$ regret bound. Intuitively, the $\mathcal{O}(\sqrt{K})$ term reflects the total reward magnitude in combinatorial semi-bandits.

We conclude this section by comparing our regret bound with existing bounds in special cases. In cascading bandits, we have $d = 0$
\footnote{As is in \citet{kveton2015cascading}, we assume that the deterministic conditional distributions in cascading bandits are known to the learning agent, thus $d=0$.}
and $C = 1$. Thus our regret bound is $\mathcal{O}(\sqrt{L \Omax n} \log n)$. On the other hand, the regret bound in \citet{kveton2015cascading} is $\mathcal{O}(\sqrt{L K n \log n})$. Since $\Omax \leq K$, our regret bound is at most $\mathcal{O}(\sqrt{\log n})$ larger. In combinatorial semi-bandits, we have $d=0$, $\Omax = K$, and $C=1$ (see Appendix~\ref{sec:assumptions}). Thus, our regret bound reduces to $R_B(n) \leq \mathcal{O} \left( \sqrt{L K n} \log n \right)$. On the other hand, Theorem 6 of \citet{pmlr-v38-kveton15} derives a $\mathcal{O} \left( \sqrt{L K n  \log n} \right)$ worst-case regret bound for a UCB-like algorithm for combinatorial semi-bandits. Our regret bound is only $\mathcal{O} \left( \sqrt{\log n} \right)$ larger. 

%% file: relatedwork.tex
\section{Related Work}

In general, it is challenging to calculate the exact posterior distribution in Thompson sampling in complex problems. \citet{urteaga2018variational} used variational inference to approximate the posterior of arms by a mixture of Gaussians. The main difference in our work is that we focus on structured arms and rewards, where the rewards are correlated through latent variables. Theoretical analysis of approximate inference in Thompson sampling was done in \citet{phan2019thompson}. By matching the minimal exploration rates of sub-optimal arms, \citet{combes2017minimal} solved a different class of structured bandit problems.

\citet{gopalan14thompson} and \citet{kawale15efficient} used particle filtering to approximate the posterior in Thompson sampling. Particle filtering is known to be consistent. However, in practice, its performance depends heavily on the number of particles. When the number of particles is small, particle filtering is computationally efficient but achieves poor approximation results. This issue can be alleviated by particle-based Bayesian sampling~\cite{zhang2020stochastic}. On the other hand, when the number of particles is large, particle filtering performs well but is computationally demanding. 

\citet{blundell2015weight} used variational inference to approximate the posterior in neural networks and incorporated it in Thompson sampling \citep{blundell2015weight}. \citet{haarnoja2017reinforcement,haarnoja2018soft} learned an energy-based policy determined by rewards, which is approximated by minimizing the KL divergence between the optimal posterior and variational distribution. 
The intractable posterior of Thompson sampling in neural networks can be approximated in the last layer, which is treated as features in Bayesian linear regression \citep{riquelme18deep,liu18customized}. However, the uncertainty may be underestimated in bandit problems~\citep{zhang2019scalable} or zero uncertainty is propagated via Bellman error~\citep{osband2018randomized}. 
This can be alleviated by particle-based Thompson sampling \citep{lu2017ensemble,zhang2019scalable}. Follow-the-perturbed-leader exploration \cite{kveton19garbage,kveton19perturbed,kveton19perturbed2,vaswani20old,kveton20randomized} is an alternative to Thompson sampling that does not require posterior.

%% file: experiments.tex
\section{Experiments}

In this section, we show that our algorithms can be applied beyond traditional models and \emph{learn more general models}. Specifically, we compare our approaches to traditional bandit algorithms for the cascade model, position-based model, and rank-$1$ bandit. The performance of the algorithms is measured by their average cumulative reward. The \emph{average cumulative reward} at time $n$ is $\frac{1}{n}\sum_{t=1}^n r(x_t, z_t)$, where $r(x_t, z_t)$ is the reward received at time $t$. The rewards of various models are defined in \cref{sec:model1,sec:model2,sec:model3}. We report the average results over $20$ runs with standard errors. 

Note that $\idTSvi$ and $\idTSivi$ are expected to perform worse than $\idTS$ since they are approximations. We now briefly discuss how to justify that $\idTSvi$ and $\idTSivi$ perform similarly to $\idTS$. Recall that due to latent variables, it is computationally intractable to implement the exact Thompson sampling $\idTS$ in general influence diagram bandits. However, we can efficiently compute an upper bound on the average cumulative reward of $\idTS$ based on numerical experiments in a feedback-relaxed setting. Specifically, consider a feedback-relaxed setting where all latent nodes $Z$ are observed. Note that $\idTS$ is computationally efficient in this new setting since it is fully observed. Moreover, due to more information feedback, $\idTS$ should perform better in this feedback-relaxed setting and hence provide an upper bound on the average cumulative reward of $\idTS$ in the original problem. In this section, we refer to $\idTS$ in this feedback-relaxed setting as $\idTSfull$, to distinguish it from $\idTS$ in the original problem. Intuitively, if $\idTSfull$ and $\idTSvi$ perform similarly, we also expect $\idTSvi$ and $\idTS$ to perform similarly. 

To speed up $\idTSvi$ and $\idTSivi$ in our experiments, we do not run the \say{while} loop in line $9$ of \cref{alg:vits} until convergence. In $\idTSvi$, we run it only once. In $\idTSivi$, which is less stable in estimating $q_t(z_t)$ but more computationally efficient, we run it up to $30$ times. We did not observe any significant drop in the quality of $\idTSvi$ and $\idTSivi$ policies when we used this approximation.

\begin{figure*}
    \centering
     \begin{subfigure}[t]{0.31\textwidth}
    \includegraphics[width=0.88\textwidth]{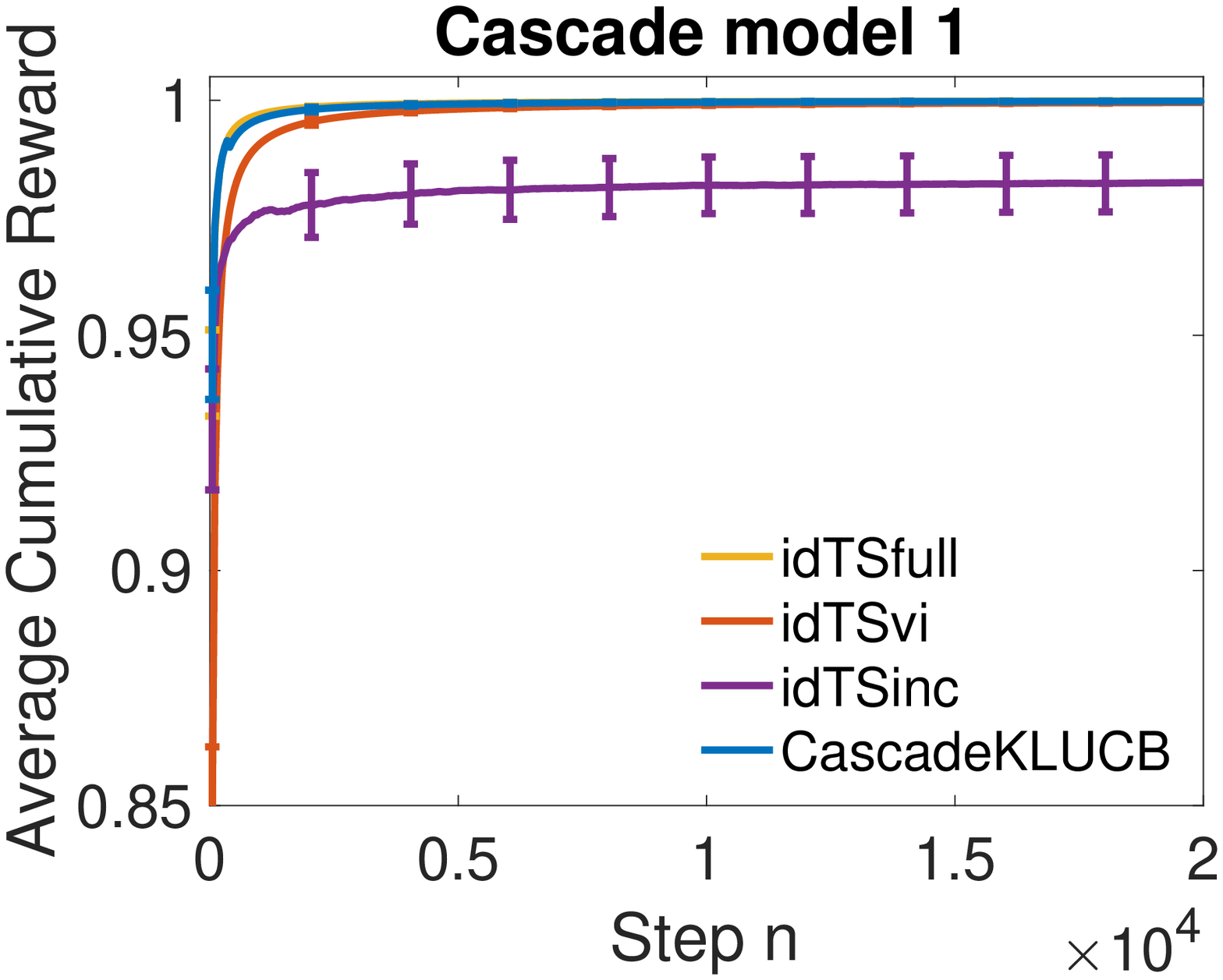}
    \caption{}
    \label{fig:res1}
     \end{subfigure}
      \begin{subfigure}[t]{0.301\textwidth}
    \includegraphics[width=0.88\textwidth]{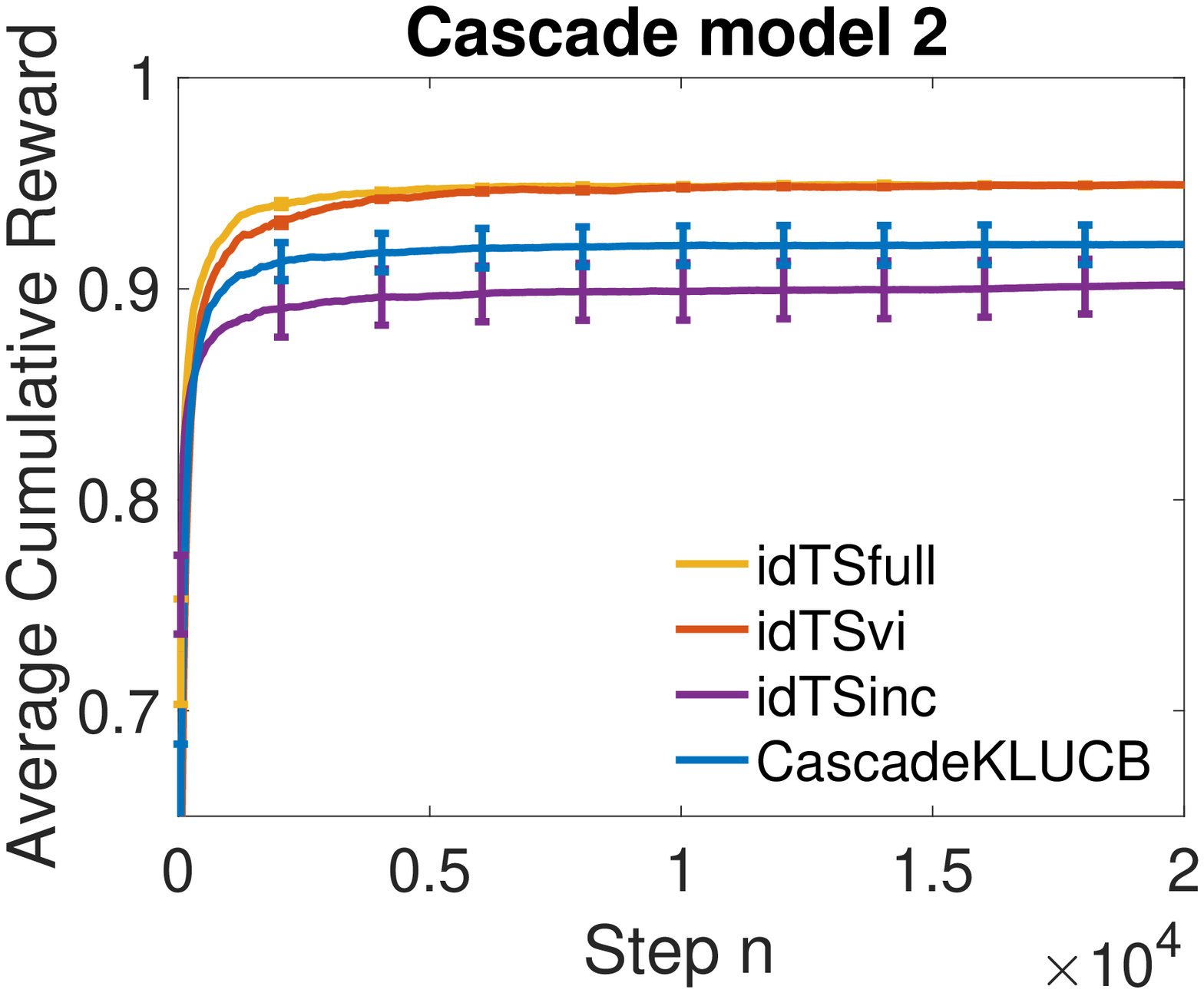}
    \caption{}
    \label{fig:res2}
     \end{subfigure}
     \begin{subfigure}[t]{0.31\textwidth}
    \includegraphics[width=0.88\textwidth]{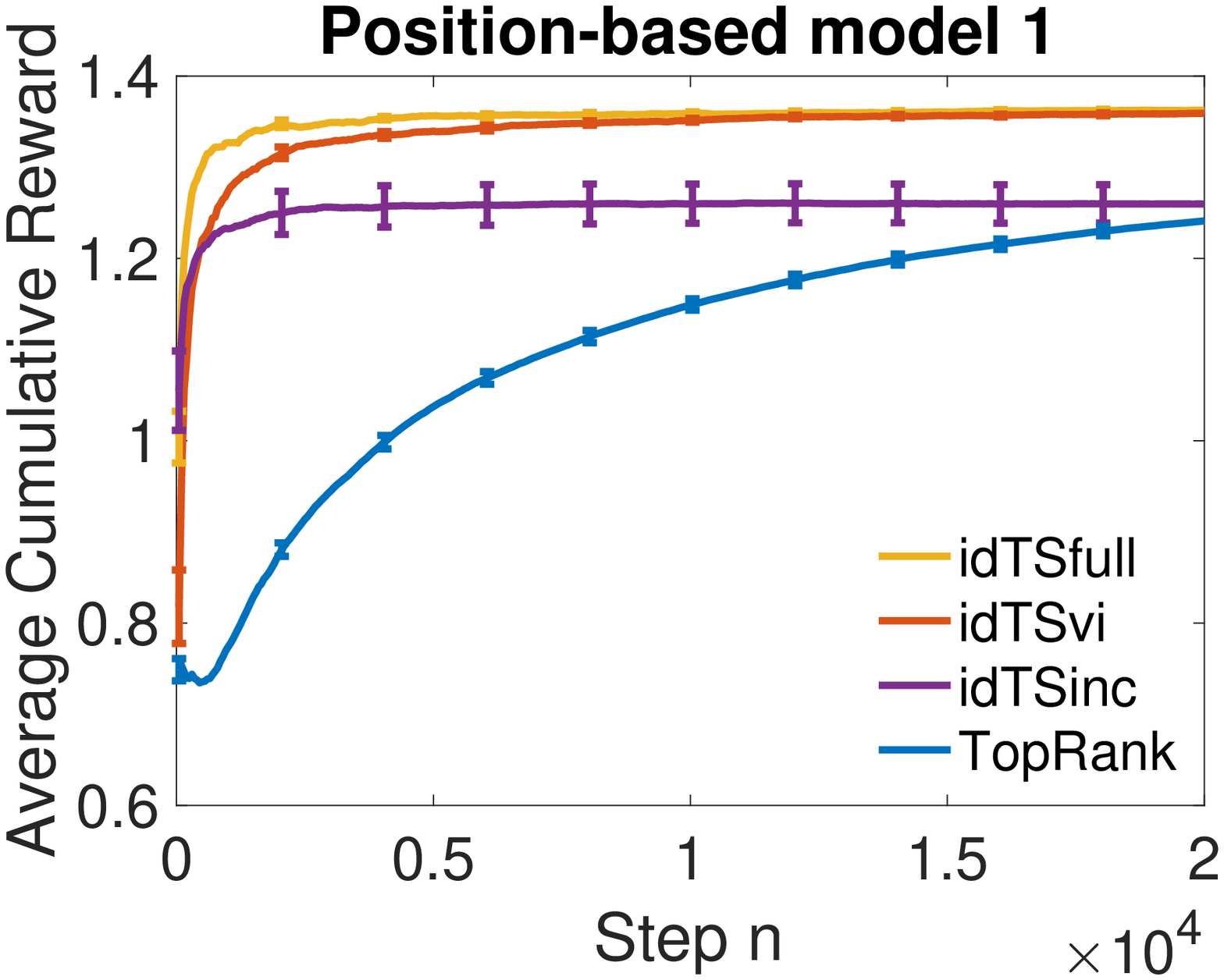}
    \caption{}
    \label{fig:res3}
     \end{subfigure} \\
     \vspace{0.2cm}
      \begin{subfigure}[t]{0.31\textwidth}
    \includegraphics[width=0.88\textwidth]{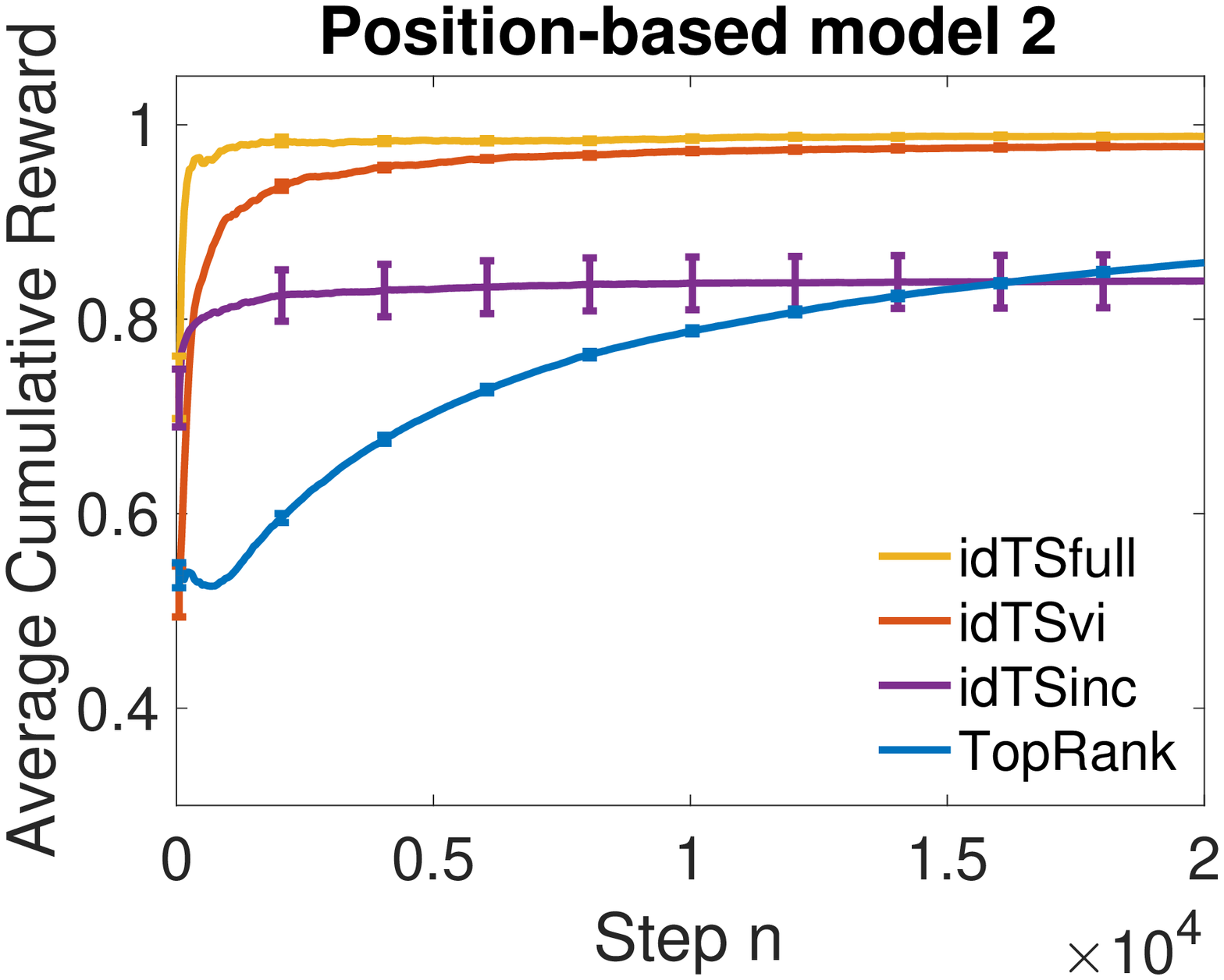}
    \caption{}
    \label{fig:res4}
     \end{subfigure}
     \begin{subfigure}[t]{0.31\textwidth}
    \includegraphics[width=0.88\textwidth]{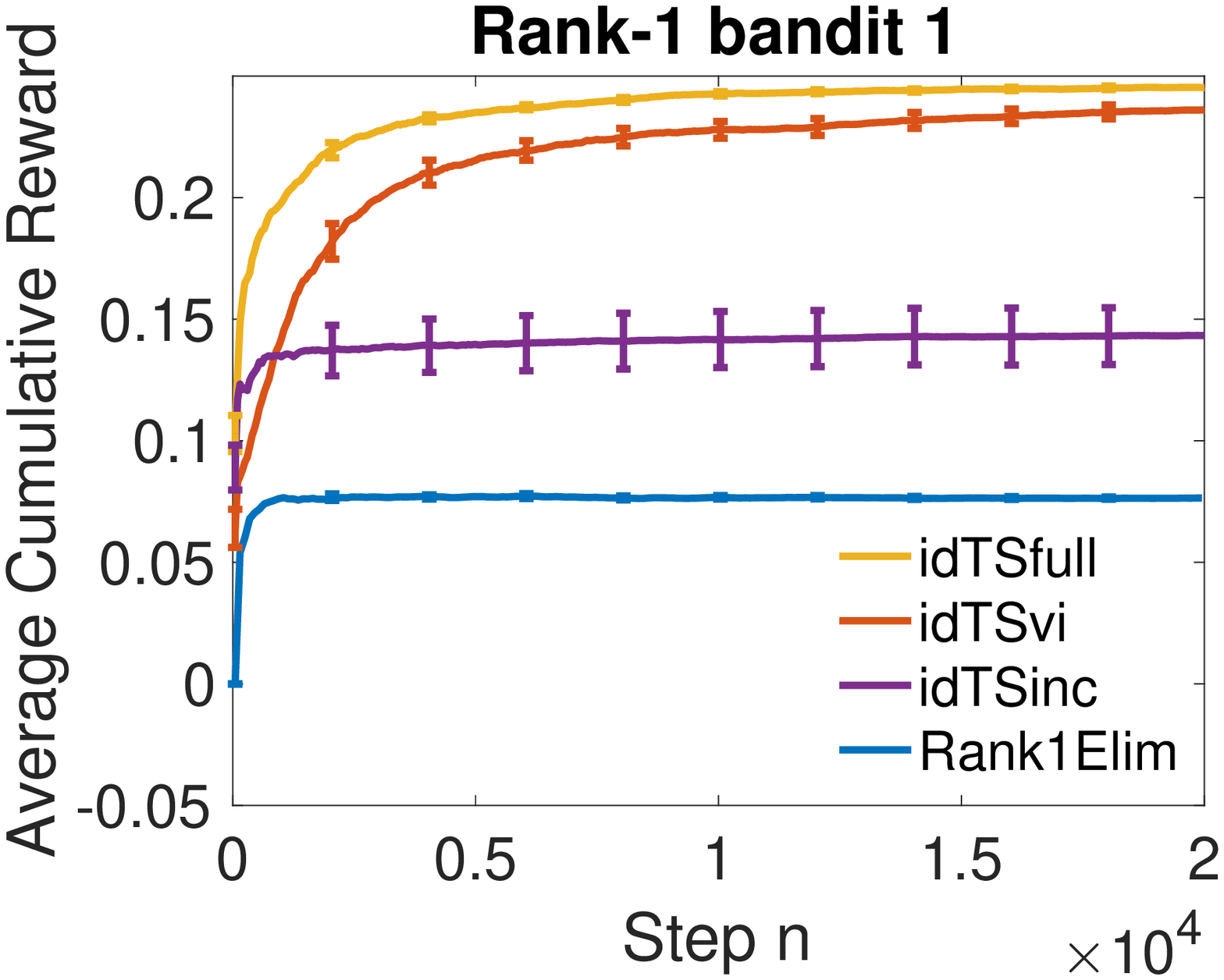}
    \caption{}
    \label{fig:res5}
     \end{subfigure}
     \begin{subfigure}[t]{0.31\textwidth}
    \includegraphics[width=0.88\textwidth]{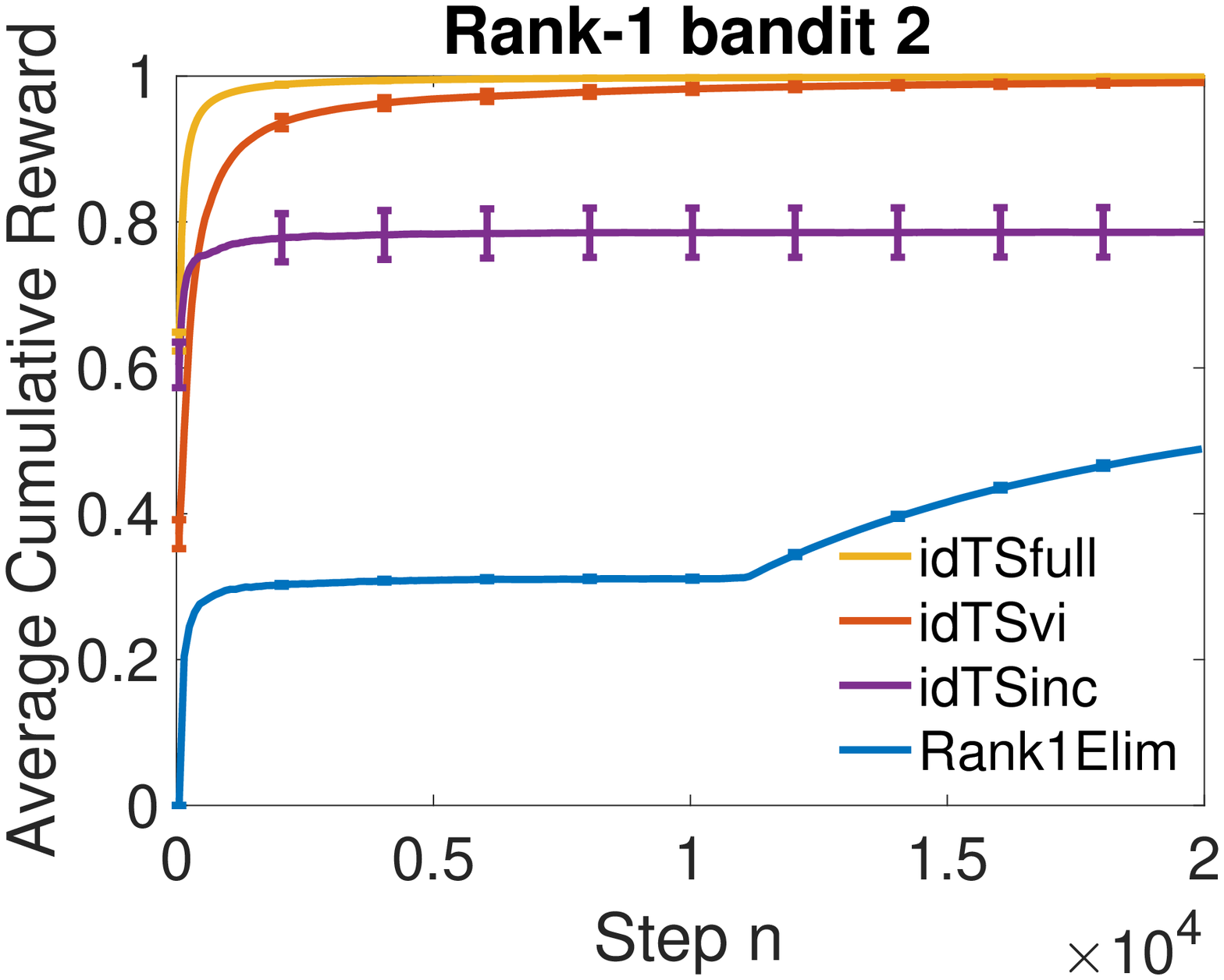}
    \caption{}
    \label{fig:res6}
     \end{subfigure}
     \caption{A comparison between different bandit algorithms in various structured bandit problems.}
    \label{fig:res}
\end{figure*}

\subsection{Cascade Model}
\label{sec:exp1}

Algorithms for the cascade model, such as $\cascadeklucb$, make strong assumptions. On the other hand, our algorithms make fewer assumptions and allow us to learn more general models. First, we experiment with the cascade model (\cref{sec:model1}). The number of items is $L = 20$ and the length of the list is $K = 2$. The attraction probability of item $i$ is $\theta_i = i / 20$. We refer to this problem as \emph{cascade model $1$}. Second, we experiment with a variant of the problem where the cascade assumption is violated, which we call \emph{cascade model 2}. In cascade model 2, we modify the conditional dependencies of $C_k$ and $E_k$ as
\begin{align*}
  P(C_k = 1 \mid W_k, E_k)
  & = (1- W_k) E_k\,, \\
  P(E_k = 1 \mid C_{k - 1}, E_{k - 1})
  & = C_{k - 1} E_{k - 1}\,.
\end{align*}
This means that an item is clicked only if it is not attractive and its position is examined, and an item is examined only if the previous item is examined and clicked. 

Our results are reported in \cref{fig:res1,fig:res2}. We observe several trends. First, among all our algorithms, $\idTSfull$ and $\idTSvi$ perform the best. $\idTSivi$ performs the worst and its runs have a lot of variance, which is caused by an inaccurate estimation of $q(z)$. Second, $\idTSivi$ is much more computationally efficient than $\idTSvi$. For each run in cascade model $1$, $\idTSvi$ needs $1990.502$ seconds on average, while its incremental version $\idTSivi$ only needs $212.772$ seconds. Third, in Figure \ref{fig:res2}, we observe a clear gap between $\cascadeklucb$ and our algorithms, $\idTSfull$ and $\idTSvi$. This is because the cascade assumption is violated, and thus $\cascadeklucb$ cannot effectively leverage the problem structure. In contrast, our algorithms are general and still effectively identify the best action.

\subsection{Position-Based Model}
\label{sec:exppbm}

We also experiment with the position-based model, which is detailed in \cref{sec:model1}. The number of items is $L = 20$ and the length of the list is $K = 2$. The attraction probability of item $i$ is $\theta_i = i / 20$. We consider two variants of the problem. In \emph{position-based model 1}, the examination probabilities of both positions are $0.7$. In \emph{position-based model 2}, the examination probabilities are $0.8$ and $0.2$. The baseline is a state-of-the-art bandit algorithm for the position-based model, $\toprank$ \citep{lattimore2018toprank}. Our results are reported in \cref{fig:res3,fig:res4}. We observe that $\idTSfull$ and $\idTSvi$ clearly outperform $\toprank$, while $\idTSivi$ achieves similar rewards compared to $\toprank$.

\subsection{Rank-$1$ Bandit}

We also evaluate our algorithms in a rank-$1$ bandit, which is detailed in Section \ref{sec:model3}. The underlying rank-$1$ matrix is $U V^T$, where $U \in [0, 1]^8$ and $V \in [0, 1]^{10}$. We consider two problems. In \emph{rank-$1$ bandit $1$}, $U_i = i / 16$ and $V_i = i / 20$. In \emph{rank-$1$ bandit $2$}, $U_i = i / 8$ and $V_i = i / 10$. The baseline is a state-of-the-art algorithm for the rank-$1$ bandit, $\rankoneelim$ \citep{katariya2017stochastic}. As shown in \cref{fig:res5,fig:res6}, our algorithms outperform $\rankoneelim$. In \cref{fig:res6}, $\rankoneelim$ improves significantly after $10000$ steps. The reason is that $\rankoneelim$ is an elimination algorithm, and thus improves sharply at the end of each elimination stage. Nevertheless, a clear gap remains between our algorithms and $\rankoneelim$. The average cumulative reward of our algorithms is up to three times higher than that of $\rankoneelim$.

\begin{figure*}[tph]
    \centering
     \begin{subfigure}[t]{0.31\textwidth}
    \includegraphics[width=0.88\textwidth]{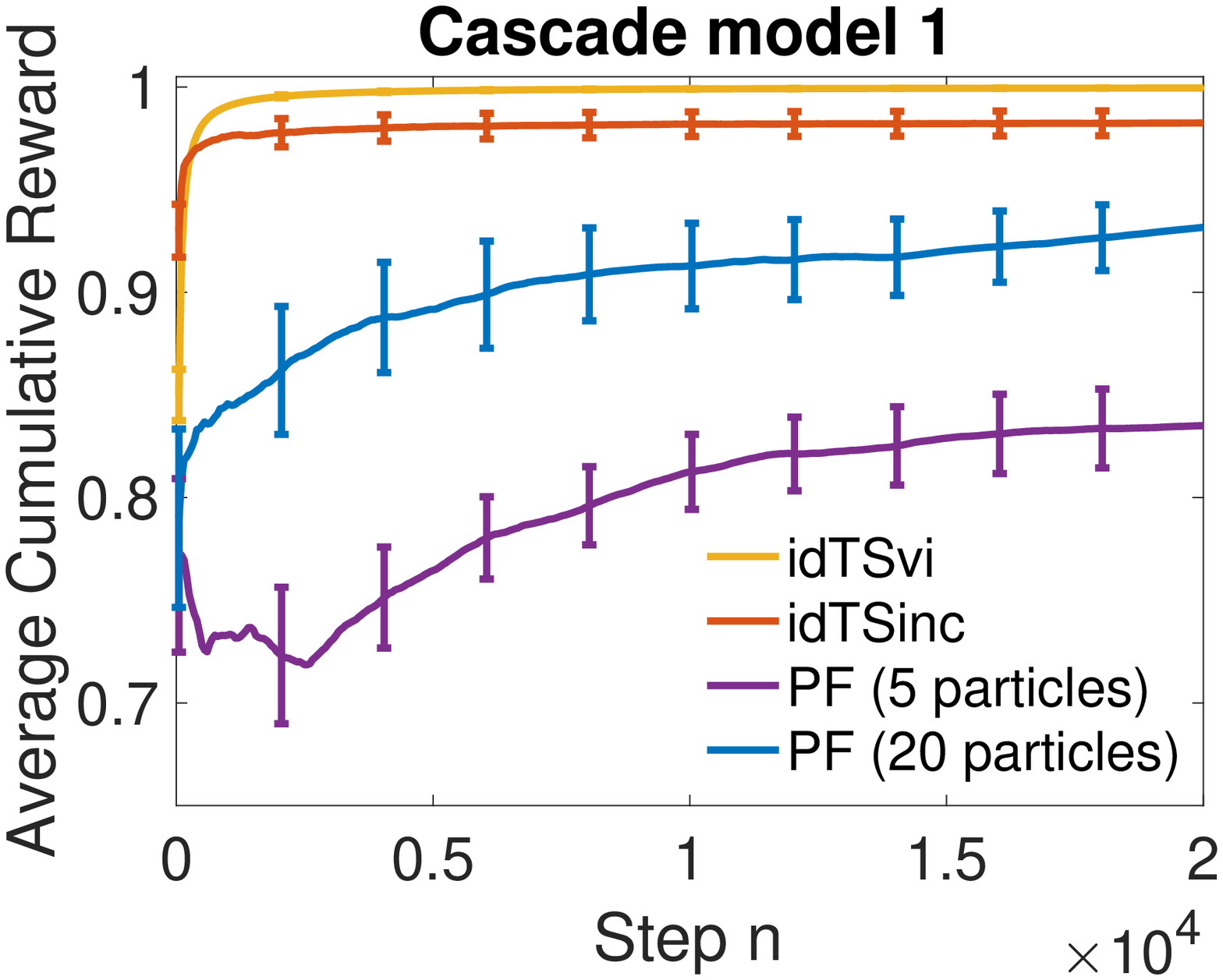}
    \caption{}
    \label{fig:respf1}
     \end{subfigure}
      \begin{subfigure}[t]{0.309\textwidth}
    \includegraphics[width=0.88\textwidth]{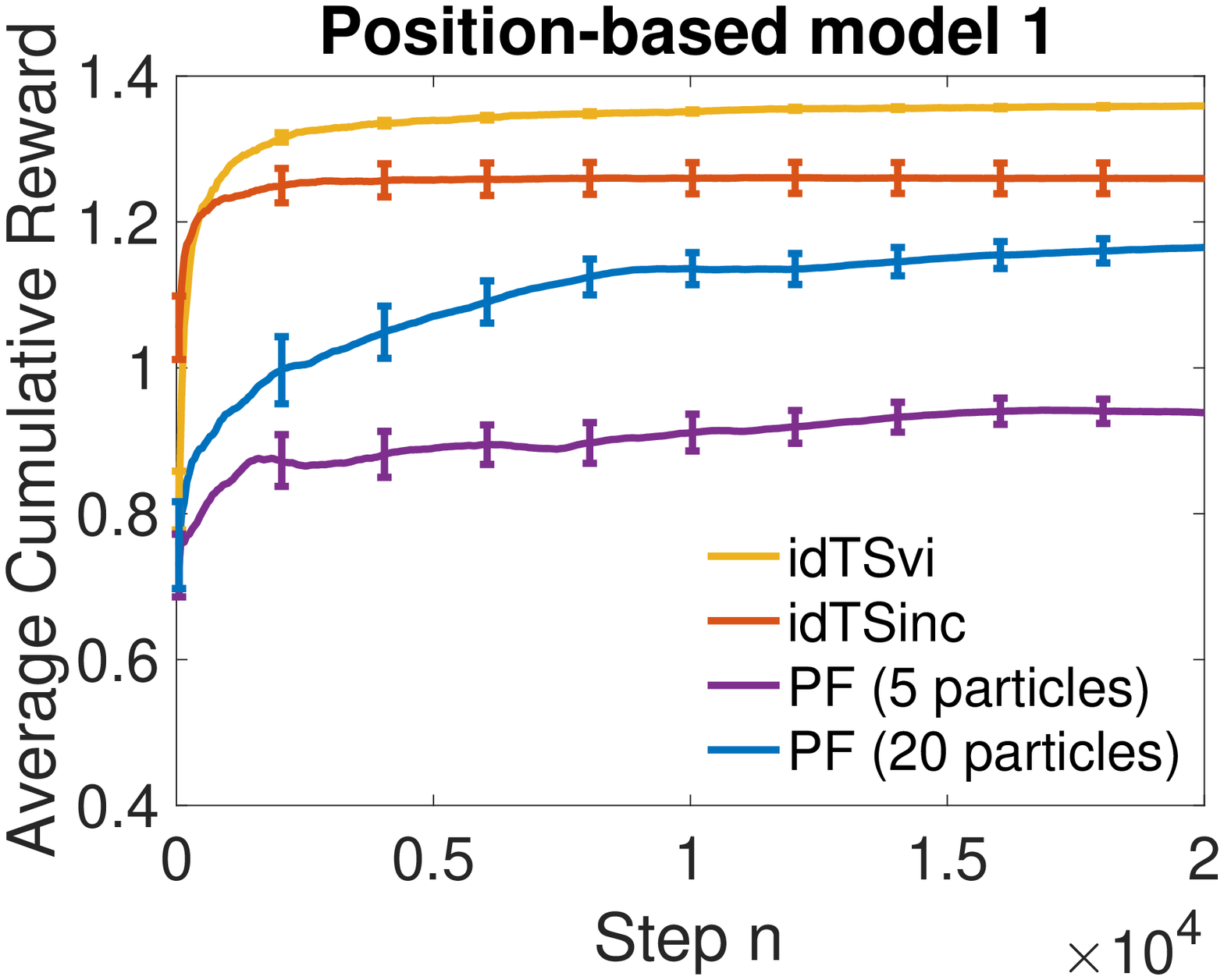}
    \caption{}
    \label{fig:respf2}
     \end{subfigure}
     \begin{subfigure}[t]{0.31\textwidth}
    \includegraphics[width=0.88\textwidth]{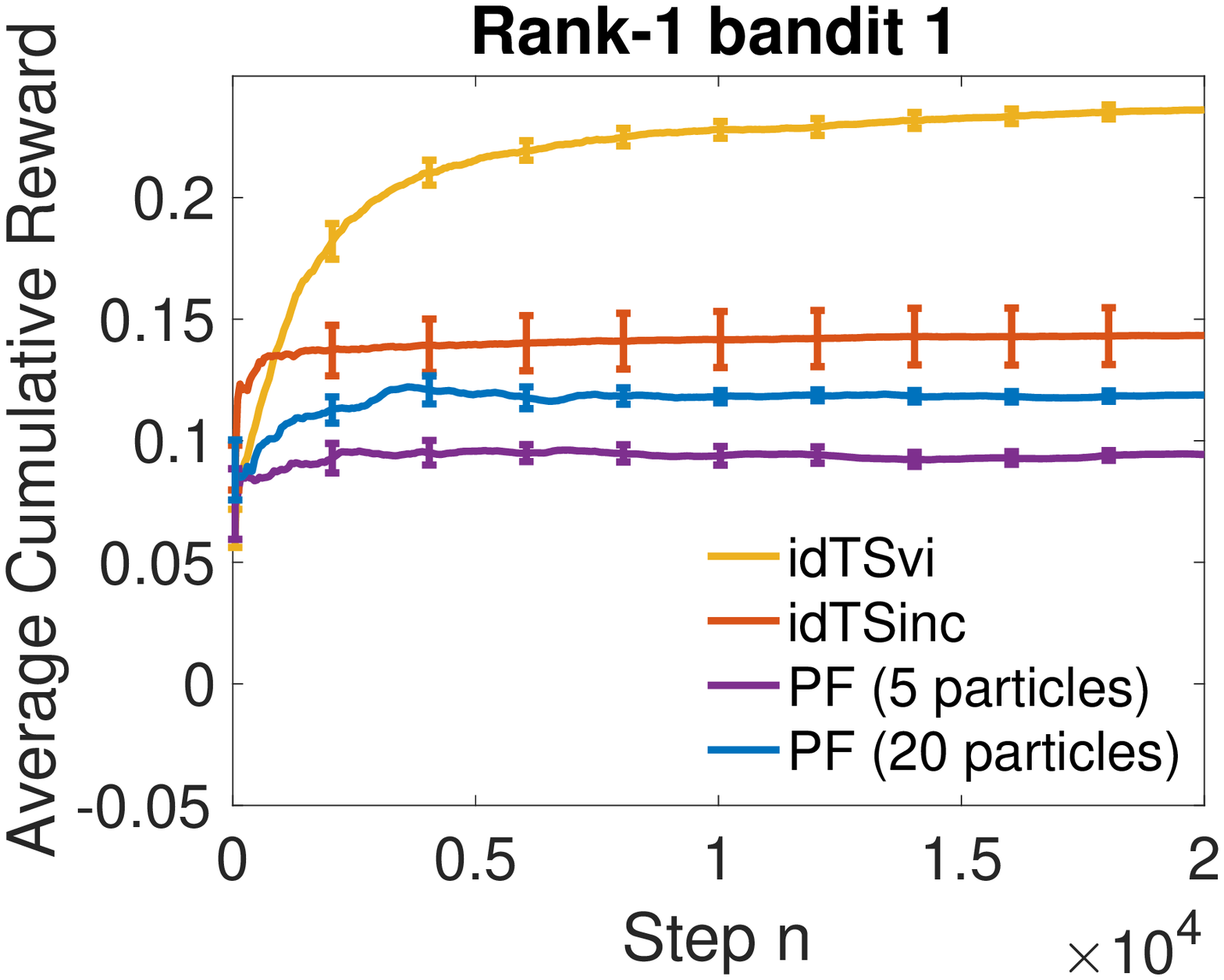}
    \caption{}
    \label{fig:respf3}
     \end{subfigure}
     \caption{A comparison between variational inference and particle filtering (PF) with different numbers of particles, when approximating the posterior of Thompson sampling in influence diagram bandits.}
     \label{fig:particle}
\end{figure*}

\subsection{Comparison to Particle Filtering}
\label{sec:com2pf}

In this section, we compare variational inference to particle filtering for posterior sampling in influence diagram bandits. As discussed in Section \ref{sec:learninginfluence}, our algorithms are based on variational inference, considering that the performance of particle filtering depends on hard to tune parameters, such as the number of particles and transition prior. We validate the advantages of variational inference in cascade model $1$, position-based model $1$, and rank-$1$ bandit $1$. 

We implement particle filtering as in \citet{andrieu2003introduction}. Let $m$ be the number of particles. At each time, the algorithm works as follows. First, we obtain $m$ models (particles) by sampling them from their Gaussian transition prior. Second, we set their importance weights based on their likelihoods. Similarly to $\idTSfull$, we assume that the latent variables are observed. This simplifies the computation of the likelihood, and gives particle filtering an unfair advantage over $\idTSvi$ and $\idTSivi$. Third, we choose the model with the highest weight, take the best action under that model, and observe its reward. Finally, we resample $m$ particles proportionally to their importance weights. The new particles serve as the mean values of the Gaussian transition prior at the next time.

Our results are reported in \cref{fig:particle}. First, we observe that variational inference based algorithms, $\idTSvi$ and $\idTSivi$, perform clearly better than particle filtering. The performance of particle filtering is unstable and has a higher variance. Second, the performance of particle filtering is sensitive to the number of particles used. By increasing the number from $5$ to $20$, the performance of particle filtering improves. Third, the performance of particle filtering may drop over time, because sometimes particles with low likelihoods are sampled, although with a small probability.

To further show the advantage of our algorithms, we compare the computational cost of variational inference and particle filtering in \cref{tab:tab}. This experiment is in cascade model $1$, and we report the run time (in seconds) and average cumulative reward at $20000$ steps. The run time is measured on a computer with one $2.9$ GHz Intel Core i$7$ processor and $16$ GB memory. We observe that $\idTSvi$ has the longest run time, while $\idTSivi$ is much faster. As we increase the number of particles, the run time of particle filtering increases, since we need to evaluate the likelihood of more particles. At the same run time, $\idTSivi$ achieves a higher reward than particle filtering.

\begin{table}[h]
    \centering
    \vspace{0.2cm}
    \tabcolsep = 0.15cm
    \begin{tabular}{l|r|r}
    \hline
       & Run time & Reward \\\hline
       $\idTSvi$ & 1990.502 $\pm$ 19.702 & 0.999 $\pm$ 0.001\\
       $\idTSivi$ & 212.772 $\pm$ 2.204 & 0.983 $\pm$ 0.006\\
       PF ($5$ particles) & 117.856 $\pm$ 1.049 & 0.835 $\pm$ 0.019 \\
       PF ($10$ particles) & 246.291 $\pm$ 3.327 & 0.878 $\pm$ 0.015 \\
       PF ($15$ particles) & 357.284 $\pm$ 2.529 & 0.926 $\pm$ 0.013 \\
       PF ($20$ particles) & 454.844 $\pm$ 3.054 & 0.932 $\pm$ 0.015 \\\hline
    \end{tabular}
    \caption{The comparison of variational inference and particle filtering (PF) in cascade model $1$. We report the run time (in seconds) and average cumulative reward at $20000$ steps. The results are averaged over $20$ runs.}
    \label{tab:tab}
\end{table}

%% file: conclusions.tex
\section{Conclusions}

Existing algorithms for structured bandits are tailored to specific models, and thus hard to extend. This paper overcomes this limitation by proposing a novel online learning framework of influence diagram bandits, which unifies and extends most existing structured bandits. 
We also develop efficient algorithms that learn to make the best decisions under influence diagrams with complex structures, latent variables, and exponentially many actions. We further derive an upper bound on the regret of our algorithm $\idTS$. Finally, we conduct experiments in various structured bandits: cascading bandits, online learning to rank in the position-based model, and rank-$1$ bandits. The experiments demonstrate that our algorithms perform well and are general. 

As we have discussed, this paper focuses on influence diagram bandits with Bernoulli random variables. We believe that the Bernoulli assumption is without loss of generality in the sense that, by appropriately modifying some technical assumptions, our developed algorithms and analysis results can be extended to more general cases with categorical or continuous random variables. We leave the rigorous derivations in these more general cases to future work.

%% file: appendix.tex
\newpage
\appendix

\onecolumn
\newpage 
\section{Discussion of Assumptions}
\label{sec:assumptions}
In this section, we prove that the combinatorial semi-bandit and the cascading bandit satisfy Assumptions \ref{ass:mono} and \ref{ass:decomp} proposed in Section~\ref{sec:regret_bound}.

\subsection{Combinatorial Semi-Bandits}
Notice that in a combinatorial semi-bandit, the action $a = (a_1, \ldots, a_K)$, and 
\[
r(a, \theta) = \sum_{k=1}^K \theta^{(a_k)} = \sum_{l=1}^L 
\theta^{(l)} \mathbf{1} \left( l \in a \right).
\]
Thus, for any $l$, $r(a, \theta)$ is weakly increasing in $\theta^{(l)}$. Hence \cref{ass:mono} is satisfied. On the other hand, we have
\begin{align}
    \left|r(a, \theta_1) - r(a, \theta_2) \right | =& \, \left | \sum_{l=1}^L \left(\theta_1^{(l)} - \theta_2^{(l)} \right) \mathbf{1} \left( l \in a \right) \right| \nonumber \\
    \leq & \, \sum_{l=1}^L \left | \theta_1^{(l)} - \theta_2^{(l)} \right | \mathbf{1} \left( l \in a \right) = \sum_{l=1}^L  P \left( E^{(l)} \middle |  \theta_2, a \right) \left | \theta_1^{(l)} - \theta_2^{(l)} \right |,
\end{align}
where the last quality follows from the fact that all nodes in a combinatorial semi-bandit is observed, and hence
$P \left( E^{(l)} \middle |  \theta, a \right) = \mathbf{1} \left( l \in a \right) $ for all $\theta$. Thus, \cref{ass:decomp} is satisfied with $C=1$.

\subsection{Cascading Bandits}
For a cascading bandit, the action is $a = \left(a_1, \ldots, a_K \right)$, and
\[
r(a, \theta) = 1 - \prod_{k=1}^K (1 - \theta^{(a_k)})=
 1 - \prod_{l \in a} (1 - \theta^{(l)}).
\]
Thus, for any $l$, $r(a, \theta)$ is weakly increasing in $\theta^{(l)}$. Hence \cref{ass:mono} is satisfied. On the other hand, from \citet{kveton2015cascading}, we have
\begin{align}
r(a, \theta_1) - r(a, \theta_2) =& \, \sum_{k=1}^K \prod_{k_1=1}^{k-1} \left(1- \theta_2^{\left(a_{k_1} \right)} \right) \left(\theta_1^{\left(a_k\right)} - \theta_2^{\left(a_k\right)} \right) \prod_{k_2 = k+1}^K 
\left(1- \theta_1^{\left(a_{k_2} \right)} \right)  \nonumber \\
=& \, 
\sum_{k=1}^K  
P \left(E^{(a_k)} \middle | \theta_2, a \right) \left(\theta_1^{\left(a_k\right)} - \theta_2^{\left(a_k\right)} \right) \prod_{k_2 = k+1}^K 
\left(1- \theta_1^{\left(a_{k_2} \right)} \right),  \nonumber
\end{align}
where the second equality follows from $P \left(E^{(a_k)} \middle | \theta_2, a \right) = \prod_{k_1=1}^{k-1} \left(1- \theta_2^{\left(a_{k_1} \right)} \right)$. Thus, we have
\begin{align}
 \left |
 r(a, \theta_1) - r(a, \theta_2)
 \right| =& \,
 \left |
 \sum_{k=1}^K  
P \left(E^{(a_k)} \middle | \theta_2, a \right) \left(\theta_1^{\left(a_k\right)} - \theta_2^{\left(a_k\right)} \right) \prod_{k_2 = k+1}^K 
\left(1- \theta_1^{\left(a_{k_2} \right)} \right)
 \right | \nonumber \\
 \leq & \,
 \sum_{k=1}^K  
P \left(E^{(a_k)} \middle | \theta_2, a \right) \left | \theta_1^{\left(a_k\right)} - \theta_2^{\left(a_k\right)} \right | \prod_{k_2 = k+1}^K 
\left(1- \theta_1^{\left(a_{k_2} \right)} \right)
 \nonumber \\
  \leq & \,
 \sum_{k=1}^K  
P \left(E^{(a_k)} \middle | \theta_2, a \right) \left | \theta_1^{\left(a_k\right)} - \theta_2^{\left(a_k\right)} \right | ,
 \nonumber 
\end{align}
where the last inequality follows from $ \prod_{k_2 = k+1}^K
\left(1- \theta_1^{\left(a_{k_2} \right)} \right) \in [0, 1]$. Thus, \cref{ass:decomp} is satisfied with $C=1$.

\section{Proof for Theorem~\ref{thm:regret_bound}}
\label{sec:proof}
\textbf{Proof:}\\
Recall that the stochastic instantaneous reward is $r(x, z)$. Note that $r(x,z)$ is bounded since its domain is finite. Without loss of generality, we assume that $r(x, z) \in [0, B]$. Thus, for any action $a$ and probability measure $\theta \in [0, 1]^{d+L}$, we have $r(a, \theta) \in [0, B]$.

Define $R_t = r(a^*, \theta_\ast)- r(a_t, \theta_\ast)$,
then by definition, we have
\begin{align}
    R_B(n) = \sum_{t=1}^n \E[R_t] = \sum_{t=1}^n \E \left[ E[R_t | \Hist_{t-1}]\right], \nonumber
\end{align}
where $\Hist_{t-1}$ is the ``history" by the end of time $t-1$, which includes all the actions and observations by that time\footnote{Rigorously speaking, $\{\Hist_t \}_{t=0}^{n-1}$ is a filtration and $\Hist_{t-1}$ is a $\sigma$-algebra.}. For any parameter index $i=1, \ldots, d+L$ and any time $t$, we define $N_{t}^{(i)} = \sum_{\tau=1}^t \mathbf{1} \left[ E_\tau^{(i)}\right]$ as the number of times that the samples corresponding to parameter $\theta^{(i)}_\ast$ have been observed by the end of time $t$,
 and $\hat{\theta}_t^{(i)}$ as the empirical mean for $\theta^{(i)}_\ast$ based on these $N_{t}^{(i)}$ observations. Then we define the upper confidence bound (UCB) $U_t^{(i)}$ and the lower confidence bound (LCB) $L_t^{(i)}$ as
\begin{align}
    U_t^{(i)} =& \, \left \{
    \begin{array}{ll}
    \min \left\{ \hat{\theta}_t^{(i)} + c \left(t, N_{t}^{(i)} \right), 1 \right\}     & \text{if $N_t^{(i)}>0$} \\
    1    & \text{otherwise}
    \end{array}
    \right . \nonumber \\
    L_t^{(i)} =& \, \left \{
    \begin{array}{ll}
    \max \left\{ \hat{\theta}_t^{(i)} - c \left(t, N_{t}^{(i)} \right), 0 \right\}     & \text{if $N_t^{(i)}>0$} \\
    0    & \text{otherwise}
    \end{array}
    \right . \nonumber 
\end{align}
where $c \left(t, N \right) = \sqrt{\frac{1.5 \log(t)}{N}}$ for any positive integer $t$ and $N$. 
Moreover, we define a probability measure $\tilde{\theta}_{t} \in [0, 1]^{d+L}$ as
\[
\vartheta_{t}^{(i)} = \left \{ \begin{array}{ll}
U_t^{(i)}     & \text{if $i \in \mathcal{I}^+$} \\
L_t^{(i)}     & \text{if $i \in \mathcal{I}^-$}
\end{array}
\right.
\]
Since both $N_{t-1}^{(i)}$ and $\hat{\theta}_{t-1}^{(i)}$ are conditionally deterministic given $\Hist_{t-1}$, and $\mathcal{I}^+$ and $\mathcal{I}^{-}$ are deterministic, 
by the definitions above, $U_{t-1}$, $L_{t-1}$ and $\vartheta_{t-1}$ are also conditionally deterministic given $\Hist_{t-1}$. Moreover, as is discussed in \citet{russo2014learning}, since we apply exact Thompson sampling $\idTS$, $\theta_\ast$ and $\theta_t$ are conditionally i.i.d. given $\Hist_{t-1}$, and $a^* = \argmax_a r(a, \theta_\ast)$ and $a_t = \argmax_a r(a, \theta_t)$. Thus, conditioning on $\Hist_{t-1}$, $r(a^*, \vartheta_{t-1})$ and $r(a_t, \vartheta_{t-1})$ are i.i.d.,
consequently, we have
\begin{align}
    \E [R_t | \Hist_{t-1}] =& \, \E [r(a^*, \theta_\ast) - r(a_t, \theta_\ast) | \Hist_{t-1}] \nonumber \\
    =& \, \E [r(a^*, \theta_\ast) - r(a^*, \vartheta_{t-1}) | \Hist_{t-1}] 
    +
    \E [r(a_t, \vartheta_{t-1}) - r(a_t, \theta_\ast) | \Hist_{t-1}]. 
\end{align}
To simplify the exposition, for any time $t$ and $i=1, \ldots, d+L$, we define
\begin{equation}
G_{t}^{(i)} =  \left \{ \left| \theta^{(i)}_\ast - \hat{\theta}_{t}^{(i)} \right | > c \left( t, N_{t}^{(i)}\right), \, N_{t}^{(i)}>0 \right \}   
= \left \{
\theta^{(i)}_\ast > U_t^{(i)} \text{ or } \theta^{(i)}_\ast < L_t^{(i)}
\right \}.
\end{equation}
Notice that $\overline{\bigcup_{i=1}^{d+L} G_t^{(i)}} = \bigcap_{i=1}^{d+L} \overline{G_t^{(i)}} = \left \{L_t \leq \theta_\ast \leq U_t \right\}$. Moreover, from \cref{ass:mono}, if $L_t \leq \theta_\ast \leq U_t$, based on the definition of $\vartheta_t$, we have
$r(a, \theta_\ast) \leq r(a, \vartheta_t)$ for all action $a$.
Thus, we have
\begin{align}
    r(a^*, \theta_\ast) - r(a^*, \vartheta_{t-1}) \stackrel{(a)}{=} & \,  \left [ r(a^*, \theta_\ast) - r(a^*, \vartheta_{t-1})
    \right ] \mathbf{1}\left(L_{t-1} \leq \theta_\ast \leq U_{t-1} \right) \nonumber \\
    + & \,
    \left [ r(a^*, \theta_\ast) - r(a^*, \vartheta_{t-1})
    \right ] \mathbf{1}\left( \bigcup_{i=1}^{d+L} G_{t-1}^{(i)}
    \right) \nonumber \\
    \stackrel{(b)}{\leq}& \, \left [ r(a^*, \theta_\ast) - r(a^*, \vartheta_{t-1})
    \right ] \mathbf{1}\left( \bigcup_{i=1}^{d+L} G_{t-1}^{(i)}
    \right)
    \nonumber \\
    \stackrel{(c)}{\leq}& \,
     B \mathbf{1}\left( \bigcup_{i=1}^{d+L} G_{t-1}^{(i)}
    \right)
     \stackrel{(d)}{\leq} 
     B
     \sum_{i=1}^{d+L}  \mathbf{1}\left( G_{t-1}^{(i)} \right), 
     \label{eqn:partial_1}
\end{align}
where equality (a) is simply a decomposition based on indicators, inequality (b) follows from the fact that $r(a, \theta_\ast) \leq r(a, \vartheta_{t-1})$ if $L_{t-1} \leq \theta_\ast \leq U_{t-1}$, inequality (c) follows from the fact that $r(X,Z) \in [0, B]$ for all $(X, Z)$ and hence $r(a, \theta) \in [0, B]$ for all $a$ and $\theta$, and inequality (d) trivially follows from the union bound of the indicators.

On the other hand, we have
\begin{align}
   r(a_t, \vartheta_{t-1}) - r(a_t, \theta_\ast) = & \, \left[r(a_t, \vartheta_{t-1}) - r(a_t, \theta_\ast) \right] \mathbf{1} \left(L_{t-1} \leq \theta_\ast \leq U_{t-1} \right) \nonumber \\
   +& \, \left[r(a_t, \vartheta_{t-1}) - r(a_t, \theta_\ast) \right] \mathbf{1} \left( 
   \bigcup_{i=1}^{d+L} G_{t-1}^{(i)}
   \right). \nonumber 
\end{align}
Similarly as the above analysis, we have
\begin{align}
   \left[r(a_t, \vartheta_{t-1}) - r(a_t, \theta_\ast) \right] \mathbf{1} \left( 
  \bigcup_{i=1}^{d+L} G_{t-1}^{(i)}
   \right) \leq B \sum_{i=1}^{d+L}
   \mathbf{1} \left( 
G_{t-1}^{(i)}
   \right).
\end{align}
On the other hand, we have
\begin{align}
    \left[r(a_t, \vartheta_{t-1}) - r(a_t, \theta_*) \right] \mathbf{1} \left(L_{t-1} \leq \theta_* \leq U_{t-1} \right) \stackrel{(a)}{\leq} & \, C \sum_{i=1}^{d+L} P \left( E^{(i)}_t \middle | \theta_*, a_t \right) \left |  \vartheta_{t-1}^{(i)} - \theta^{(i)}_* \right | \mathbf{1} \left(L_{t-1} \leq \theta_* \leq U_{t-1} \right) \nonumber \\
    \stackrel{(b)}{\leq} & \, C \sum_{i=1}^{d+L} P \left( E^{(i)}_t \middle | \theta_*, a_t \right) \left[  U_{t-1}^{(i)} - L_{t-1}^{(i)} \right] \mathbf{1} \left(L_{t-1} \leq \theta_* \leq U_{t-1} \right) \nonumber \\
    \stackrel{(c)}{\leq} & \, C \sum_{i=1}^{d+L} P \left( E^{(i)}_t \middle | \theta_*, a_t \right) \left[  U_{t-1}^{(i)} - L_{t-1}^{(i)} \right],  \nonumber
\end{align}
where inequality (a) follows from \cref{ass:decomp}, inequality (b) follows trivially from $L_{t-1} \leq \theta_* \leq U_{t-1}$ and the definition of $\vartheta_{t-1}$, and inequality (c) follows from the fact that $U_{t-1}^{(i)} > L_{t-1}^{(i)}$ always holds, no matter what $\theta_*$ is. 
Combining the above results, we have
\begin{align}
     \E [R_t | \Hist_{t-1}]  \leq & \, C \sum_{i=1}^{d+L} \E \left[ P \left( E^{(i)}_t \middle | \theta_*, a_t \right) \left[  U_{t-1}^{(i)} - L_{t-1}^{(i)} \right] \middle | \Hist_{t-1} \right] + 2B \sum_{i=1}^{d+L} \E \left[\mathbf{1}\left(G_{t-1}^{(i)} \right) \middle | \Hist_{t-1}  \right] \nonumber \\
     \stackrel{(a)} = & \, 
      C \sum_{i=1}^{d+L} \E \left[ P \left( E^{(i)}_t \middle | \theta_*, a_t \right) \middle | \Hist_{t-1} \right] \left[  U_{t-1}^{(i)} - L_{t-1}^{(i)} \right]  + 2B \sum_{i=1}^{d+L} \E \left[\mathbf{1}\left(G_{t-1}^{(i)} \right) \middle | \Hist_{t-1}  \right] \nonumber \\
      \stackrel{(b)} = & \, 
      C \sum_{i=1}^{d+L} \E \left[ \E \left[ \mathbf{1} \left( E^{(i)}_t \right) \middle | \theta_*, a_t \right ] \middle | \Hist_{t-1} \right] \left[  U_{t-1}^{(i)} - L_{t-1}^{(i)} \right]  + 2B \sum_{i=1}^{d+L} \E \left[\mathbf{1}\left(G_{t-1}^{(i)} \right) \middle | \Hist_{t-1}  \right] \nonumber \\
    \stackrel{(c)} = & \, 
      C \sum_{i=1}^{d+L} \E \left[ \E \left[ \mathbf{1} \left( E^{(i)}_t \right) \middle | \theta_*, a_t, \Hist_{t-1} \right ] \middle | \Hist_{t-1} \right] \left[  U_{t-1}^{(i)} - L_{t-1}^{(i)} \right]  + 2B \sum_{i=1}^{d+L} \E \left[\mathbf{1}\left(G_{t-1}^{(i)} \right) \middle | \Hist_{t-1}  \right] \nonumber \\
    \stackrel{(d)} = & \, 
      C \sum_{i=1}^{d+L} \E \left[  \mathbf{1} \left( E^{(i)}_t \right) \left[  U_{t-1}^{(i)} - L_{t-1}^{(i)} \right]  \middle | \Hist_{t-1} \right]   + 2B \sum_{i=1}^{d+L} \E \left[\mathbf{1}\left(G_{t-1}^{(i)} \right) \middle | \Hist_{t-1}  \right], \nonumber
\end{align}
where (a) follows from the fact that $U_{t-1}$ and $L_{t-1}$ are deterministic conditioning on $\Hist_{t-1}$, (b) follows from the definition of $P \left( E^{(i)}_t \middle | \theta_*, a_t \right)$, (c) follows from that fact that conditioning on $\theta_*$ and $a_t$, $E_t^{(i)}$ is independent of $\Hist_{t-1}$, and (d) follows from the tower property. Thus we have
\begin{equation}
R_B(n) \leq C \E \left[\sum_{i=1}^{d+L} \sum_{t=1}^n 
\mathbf{1} \left( E^{(i)}_t \right) \left[  U_{t-1}^{(i)} - L_{t-1}^{(i)} \right] \right] + 2B \sum_{i=1}^{d+L} \sum_{t=1}^n P \left(G_{t-1}^{(i)} \right).
\label{eqn:decomp}
\end{equation}
We first bound the second term. Notice that we have
$P \left(G_{t-1}^{(i)} \right) = \E \left[ P \left(G_{t-1}^{(i)} \middle | \theta_* \right) \right]$.
For any $\theta_*$, we have
\begin{align}
    P\left(G_{t-1}^{(i)} \middle | \theta_* \right) = P \left(  \left| \theta^{(i)}_* - \hat{\theta}_{ N_{t-1}^{(i)}}^{(i)} \right | > c \left( t, N_{t-1}^{(i)}\right), \, N_{t-1}^{(i)}>0 \middle | \theta_* \right),  \nonumber
\end{align}
where we use subscript $N_{t-1}^{(i)}$ for $\hat{\theta}$ to emphasize it is an empirical mean over  $N_{t-1}^{(i)}$ samples. Following the union bound developed in \citet{auer2002finite}, we have
\begin{align}
    P\left(G_{t-1}^{(i)} \middle | \theta_* \right) =& \, P \left(  \left| \theta^{(i)}_* - \hat{\theta}_{ N_{t-1}^{(i)}}^{(i)} \right | > c \left( t, N_{t-1}^{(i)}\right), \, N_{t-1}^{(i)}>0 \middle | \theta_* \right)  \nonumber \\
    \stackrel{(a)}{\leq} & \, \sum_{N=1}^{t-1} P \left(  \left| \theta^{(i)}_* - \hat{\theta}_N^{(i)} \right | > c \left( t, N \right)  \middle | \theta_* \right)  \stackrel{(b)}{\leq} \sum_{t=1}^{N-1} \frac{2}{t^3} < \frac{2}{t^2},
    \nonumber
\end{align}
where inequality (a) follows from the union bound over the realization of $N_{t-1}^{(i)}$, and inequality (b) follows from the Hoeffding's inequality. Since the above inequality holds for any $\theta_*$, we have $P \left(G_{t-1}^{(i)} \right) < \frac{2}{t^2}$. Thus, 
\[
\sum_{i=1}^{d+L} \sum_{t=1}^n P \left(G_{t-1}^{(i)} \right) < \sum_{i=1}^{d+L} \sum_{t=1}^n \frac{2}{t^2} < (d+L) \sum_{t=1}^{\infty} \frac{2}{t^2} = \frac{(d+L) \pi^2}{3}.
\]
We now try to bound the first term of equation~\ref{eqn:decomp}. Notice that trivially, we have
\begin{align}
U^{(i)}_{t-1} - L^{(i)}_{t-1} \leq & \, 2 c \left(t, N_{t-1}^{(i)} \right) \mathbf{1}\left(N_{t-1}^{(i)} >0 \right) + \mathbf{1} \left(N_{t-1}^{(i)} = 0 \right) \nonumber \\
=& \, 2 \sqrt{\frac{1.5 \log(t)}{N_{t-1}^{(i)}}} \mathbf{1}\left(N_{t-1}^{(i)} >0 \right) + \mathbf{1} \left(N_{t-1}^{(i)} = 0 \right) \nonumber \\
\leq & \,  \sqrt{6 \log(n)} \frac{1}{\sqrt{N_{t-1}^{(i)}}}\mathbf{1}\left(N_{t-1}^{(i)} >0 \right) + \mathbf{1} \left(N_{t-1}^{(i)} = 0 \right). \nonumber 
\end{align}
Thus, we have
\begin{align}
    \sum_{i=1}^{d+L} \sum_{t=1}^n 
\mathbf{1} \left( E^{(i)}_t \right) \left[  U_{t-1}^{(i)} - L_{t-1}^{(i)} \right] \leq & \,  \sqrt{6 \log(n)} \sum_{i=1}^{d+L} \sum_{t=1}^n \frac{1}{\sqrt{N_{t-1}^{(i)}}}\mathbf{1}\left(E_t^{(i)}, \, N_{t-1}^{(i)} >0 \right) + (d+L). \nonumber
\end{align}
Notice that from the Cauchy–Schwarz inequality, we have
\begin{align}
    \sum_{i=1}^{d+L} \sum_{t=1}^n \frac{1}{\sqrt{N_{t-1}^{(i)}}}\mathbf{1}\left(E_t^{(i)}, \, N_{t-1}^{(i)} >0 \right) \leq \sqrt{\sum_{t=1}^n \sum_{i=1}^{d+L}  \mathbf{1} \left( E_t^{(i)} \right)} \sqrt{\sum_{i=1}^{d+L} \sum_{t=1}^n \frac{1}{N_{t-1}^{(i)}} \mathbf{1}\left( N_{t-1}^{(i)} >0 \right)}.
\end{align}
Moreover, we have
\[
\sum_{i=1}^{d+L} \sum_{t=1}^n \frac{1}{N_{t-1}^{(i)}} \mathbf{1}\left( N_{t-1}^{(i)} >0 \right) < (d+L) \sum_{N=1}^n \frac{1}{N} < (d+L) \left(1 + \int_{z=1}^n \frac{1}{z} dz \right) = (d+L) (1 + \log(n)).
\]
Consequently, we have
\[
\E \left[\sum_{i=1}^{d+L} \sum_{t=1}^n 
\mathbf{1} \left( E^{(i)}_t \right) \left[  U_{t-1}^{(i)} - L_{t-1}^{(i)} \right] \right] \leq \sqrt{6(d+L)  \log(n) \left(1+ \log(n) \right)} 
\E \left[
\sqrt{\sum_{t=1}^n \sum_{i=1}^{d+L}  \mathbf{1} \left( E_t^{(i)} \right)}
\right ]
+ (d+L).
\]
Moreover, we have
\begin{align}
\E \left[
\sqrt{\sum_{t=1}^n \sum_{i=1}^{d+L}  \mathbf{1} \left( E_t^{(i)} \right)}
\right ] \leq & \,
\sqrt{\sum_{t=1}^n \E \left[ \sum_{i=1}^{d+L}  \mathbf{1} \left( E_t^{(i)} \right) \right ]} 
\stackrel{(a)}{=}  \,
\sqrt{\sum_{t=1}^n \E \left[ \E \left[ \sum_{i=1}^{d+L}  \mathbf{1} \left( E_t^{(i)} \right) \middle | a_t \right ] \right ]} \nonumber \\
\leq & \,
\sqrt{\sum_{t=1}^n \E \left[ \max_a \E \left[ \sum_{i=1}^{d+L}  \mathbf{1} \left( E_t^{(i)} \right) \middle | a \right ] \right ]}
\stackrel{(b)}{=}  \,
\sqrt{\sum_{t=1}^n \E \left[ \Omax \right ]} = \, \sqrt{n \Omax},
\end{align}
where equality (a) follows from the tower property, and equality (b) follows from the definition of $\Omax$. Thus, we have
\[
\sum_{i=1}^{d+L} \sum_{t=1}^n 
\mathbf{1} \left( E^{(i)}_t \right) \left[  U_{t-1}^{(i)} - L_{t-1}^{(i)} \right] \leq \sqrt{6(d+L) \Omax n \log(n) \left(1+ \log(n) \right)} + (d+L)
\]
Putting everything together, we have
\begin{align}
    R_B(n) \leq & \, C \sqrt{6(d+L) \Omax n \log(n) \left(1+ \log(n) \right)} + \left(C+ \frac{2 \pi^2}{3} B \right)(d+L)  \nonumber \\
    =& \, \mathcal{O} \left( C \sqrt{(d+L) \Omax n} \log(n) \right) .
\end{align}
\textbf{q.e.d.}

\section{Pseudocode of $\idTSivi$}
\label{sec:idTSivi_app}
The pseudocode of $\idTSivi$ is summarized in Algorithm \ref{alg:ivits}.

\begin{algorithm}[h]
	\caption{$\idTSivi$: A computationally efficient variant of $\idTSvi$.}
	\label{alg:ivits}
	
	\begin{algorithmic}[1]
		\STATE {\bfseries Input:} $\epsilon >0$
		\STATE Randomly initialize $q$  
		\FOR{$t = 1, \dots, n$}
		\STATE Sample $\theta_t$ proportionally to $q(\theta_t)$
		\STATE Take action $a_t = \argmax_{a \in \cA^K} r(a, \theta_t)$
		
		\STATE Observes $x_t$ and receive reward $r(x_t, z_t)$
		
		\STATE Randomly initialize $q$
		
		\STATE Calculate $\mathcal{L}(q)$ using (\ref{eq:lq}) and set $\mathcal{L}'(q) = - \infty$
		\WHILE{$\mathcal{L}(q) - \mathcal{L}'(q) \geq \epsilon$}
		\STATE 	Set $\mathcal{L}'(q) = \mathcal{L}(q)$
		\STATE  Update $q_t (z_t)$ using (\ref{eq:estep}), for all $z_t$ 
		\STATE 	Update $q(\theta)$ using (\ref{eq:mstep})  
		\STATE 	Update $\mathcal{L}(q)$ using (\ref{eq:lq})
		\ENDWHILE
		\ENDFOR
	\end{algorithmic}
\end{algorithm}